\documentclass[11pt]{article}

\usepackage[preprint]{acl}

\usepackage{times}
\usepackage{latexsym}

\usepackage[T1]{fontenc}

\usepackage[utf8]{inputenc}

\usepackage{microtype}

\usepackage{inconsolata}

\usepackage{graphicx}

\title{Scaling Multilingual Mixture-of-Experts: How Language Coverage Changes Capacity and Compute Allocation}

\newcommand{\jd}[1]{\todo[inline,color=orange!40]{JD: #1}}

\newcommand{\ns}[1]{\todo[inline,color=teal]{NS: #1}} 

\author{
\bf Joel Dag\thanks{\ \ Equal contribution.}, Nikit Srivastava\footnotemark[1], Hamada M. Zahera, Axel-Cyrille Ngonga Ngomo 
\\ \\ Data Science Group, Heinz Nixdorf Institute, Paderborn University, Germany  
\\ \\ 
\{joeldag, nikit.srivastava, hamada.zahera, axel.ngonga\}@mail.uni-paderborn.de \\
}

\usepackage{amsmath}
\usepackage{amssymb}
\usepackage{booktabs}
\usepackage{colortbl}
\usepackage{float}
\usepackage[disable]{todonotes}
\usepackage[inline]{enumitem}

\begin{document}
\maketitle

\begin{abstract}
Multilingual models share capacity across languages, so adding languages can hurt performance on existing ones. Mixture-of-experts models expand this capacity with little additional per-token computation. Given this extra capacity, does adding experts reduce what each additional language costs? To answer this question, we propose a repetition-aware joint scaling law relating number of active parameters, training tokens per language, number of experts, and language coverage. Fitted on 26,300 per-language loss observations from 64 pretrained configurations, our law reduces prediction error on held-out configurations by 28\% relative to the strongest evaluated baseline. At matched number of active parameters and training tokens per language, 32 experts reduce the loss penalty of expanding from 12 to 200 languages by 83\% compared with a dense model. Our law further shows that broader language coverage favors more experts across a wider range of compute budgets, guiding capacity allocation as multilingual models scale.
\end{abstract}
\section{Introduction}
\label{sec:introduction}
Multilingual pretraining allows one model to support language understanding and generation across many languages \citep{conneau_2020_xlmr,xue_2021_mt5}.
Through cross-lingual transfer, languages with limited training data can benefit from sharing a model with other languages.
However, these benefits are uneven and depend on the amount of data, the languages combined, and the capacity of the model \citep{chang_2024_multilinguality_curse}.
Extending language coverage therefore requires understanding what each language gains or loses, not only how the average changes.

A central difficulty is the \emph{curse of multilinguality}, where adding languages worsens performance on those already supported \citep{conneau_2020_xlmr}.
Increasing model capacity can alleviate this penalty, but enlarging a dense model also increases computation for every token.
An alternative is to dedicate capacity to individual languages. 
X-MOD adds language-specific modules, while X-ELM trains separate models on subsets of languages \citep{pfeiffer_2022_lifting_curse,blevins_2024_breaking_curse}.
Both reduce competition between languages, but tie capacity to languages chosen in advance.

Sparse mixture-of-experts (MoE) models offer a different way to share capacity.
They store many experts while activating only a small subset for each token, allowing capacity to grow with approximately fixed per-token computation \citep{fedus_2022_switch}.
Multilingual models such as mSwitch and LOLA demonstrate the promise of learned expert routing across many languages \citep{fedus_2022_switch,srivastava_2025_lola}.
However, these successes leave open whether experts reduce the loss increase from adding languages, or improve performance similarly regardless of language coverage.

Scaling laws could answer this question before training, but existing laws leave the interaction between expert capacity and language coverage unresolved.
Multilingual laws study language sampling, transfer, and the cost of increasing coverage \citep{he_2025_multilingual_scaling,longpre_2026_atlas}.
In contrast, MoE laws study how to allocate capacity between active parameters and experts under compute and memory limits \citep{ludziejewski_2025_joint_moe}.
We close this gap with a joint scaling law over the number of active parameters, training tokens per language, the number of experts, and language coverage.
Many of the added languages have little unique text, so the law also accounts for repeated passes over the same data \citep{muennighoff_2023_data_constrained}.
Once fitted, the law lets us study capacity allocation beyond the training grid.

\paragraph{Research questions.}
With this law, we ask three questions about the role of expert capacity in multilingual pretraining:
\begin{enumerate}[leftmargin=*,itemsep=1pt,topsep=2pt,parsep=0pt]
    \item \textit{Does adding experts reduce what each additional language costs?}
    \item \textit{How does broader language coverage change the preferred allocation between active and expert capacity under fixed resource budgets?}
    \item \textit{When adding languages preserves average quality, how does this affect individual languages?}
\end{enumerate}

To fit the law, we pretrain 64 model configurations that independently vary the number of active parameters, the number of experts, and language coverage, yielding 26,300 per-language loss observations.
Another 37 runs cover seed replication, alternative language mixtures, and a larger-scale stress test.
We evaluate 617 checkpoints, primarily through language-modeling loss and corroborated on 14 multilingual benchmarks.
Following X-MOD, we compare models trained on different numbers of languages at matched training tokens per language so that original languages do not receive less data \citep{pfeiffer_2022_lifting_curse}.
On held-out configurations, the law reduces prediction error by 28\% relative to the strongest evaluated baseline, JointMoE \citep{ludziejewski_2025_joint_moe} (Section~\ref{sec:result-selection}).
For RQ1, 32 experts reduce the loss increase of expanding from 12 to 200 languages by 83\% relative to a dense model, a direction that holds across seeds (Section~\ref{sec:result-pressure}).
For RQ2, the law indicates that broader language coverage extends the compute range over which expert-heavy configurations are preferred (Section~\ref{sec:result-allocation}).
For RQ3, restoring average loss after expansion still leaves many original languages worse off in a representative setting, so an average alone is an insufficient target (Section~\ref{sec:result-preservation}).

Our main contributions are
\begin{enumerate*}[label=(\roman*)]
    \item a repetition-aware joint scaling law for multilingual MoEs, validated on held-out configurations;
    \item a controlled study of how expert capacity changes the cost of adding languages at matched training tokens per language; and
    \item an analysis of how coverage shifts the preferred use of experts and why average quality does not protect every language.
\end{enumerate*}
All training and analysis code is open-source and publicly available.\footnote{\url{https://github.com/dice-group/multilingual-moe-scaling-law}}
\section{Related Work}
\label{sec:related-work}
Scaling laws predict language-modeling loss from model size and training data.
\citet{hestness_2017_deep_learning_scaling} observed such power laws across deep-learning domains before \citet{kaplan_2020_scaling_laws} established them for language models.
\citet{hoffmann_2022_compute_optimal} later found that model size and training data should scale approximately equally with compute.
Other laws couple the size and data terms instead of adding them \citep{rosenfeld2020a,li2025predictable}.
\citet{muennighoff_2023_data_constrained} extend these laws to repeated data, finding that a few epochs are nearly as valuable as unique data.
MoE models add the number of experts as a further dimension, which grows the total number of stored parameters at approximately fixed per-token computation.
\citet{clark_2022_unified_scaling} relate loss to the number of active parameters and the number of experts, while \citet{ludziejewski_2024_fine_grained_moe} add expert granularity.
We build on JointMoE \citep{ludziejewski_2025_joint_moe}, which models the number of active parameters, training data, and the number of experts together.
All these laws, however, treat training data as a single distribution, which leaves language coverage outside the model.

In multilingual pretraining, languages compete for finite capacity, which trades positive cross-lingual transfer against interference \citep{conneau_2020_xlmr,arivazhagan_2019_massively_multilingual}.
Language-specific capacity reduces this interference, as in the modules of X-MOD \citep{pfeiffer_2022_lifting_curse} or the separate per-subset models of X-ELM \citep{blevins_2024_breaking_curse}.
Sparse MoE models such as GShard \citep{lepikhin_2021_gshard}, Switch Transformer \citep{fedus_2022_switch}, and NLLB \citep{nllb_team_2022_no_language_left_behind} instead learn to share capacity across languages.
Their routing reflects linguistic structure \citep{srivastava_2025_lola}, specializing by language in early and late layers while sharing across languages in middle layers \citep{bandarkar_2026_multilingual_routing}.
Other methods allocate experts by language similarity \citep{li_2025_group_then_scale,zhang_2025_less_better} or align routing across languages \citep{dong_2026_sara}.
These studies show that added capacity helps multilingual models, but none measures how the loss increase from adding languages changes with the number of experts.

Multilingual scaling laws relate loss to sampling ratios, resource levels, linguistic similarity, data size, and model capacity \citep{fernandes_2023_multilingual_scaling,chen_2023_pareto_front,chang_2024_multilinguality_curse}.
\citet{he_2025_multilingual_scaling} model how language sampling affects scaling, while \citet{cao_2026_shapleylaw} estimate each language's contribution to optimize mixtures.
ATLAS \citep{longpre_2026_atlas} further separates training tokens per language, cross-lingual transfer, repetition, and the capacity required by broader coverage.
We adapt these laws as baselines, but none studies how they change with expert capacity.
They therefore cannot indicate how to split capacity between active parameters and experts as language coverage grows.
Because multilingual benefits are uneven across languages \citep{chang_2024_multilinguality_curse}, a joint law must also predict per-language loss rather than only the average.
We address this gap with a joint law over the number of active parameters, training tokens per language, the number of experts, and language coverage. 

\section{Scaling Law Formulation}
\label{sec:scaling-law}

Given that no existing scaling law relates the loss of each language to both the number of experts and language coverage, 
we first define this prediction target (Section~\ref{sec:problem-setting}).
We then combine MoE with multilingual scaling into a core joint law (Section~\ref{sec:core-law}) and refine this law for prediction (Section~\ref{sec:refinements}).

\subsection{Problem Setting}
\label{sec:problem-setting}
Under a fixed training-token budget, broader language coverage reduces training tokens available to each language, while more languages compete for the same model capacity.
Additional experts expand this capacity without increasing the number of active parameters, which may offset the cost of broader coverage.
Measuring this offset requires a law that predicts the loss of each language from both the number of experts and language coverage.

Formally, following the notation of \citet{clark_2022_unified_scaling}, a training configuration fixes the number of active parameters $N$, the number of experts $E$, and the number of training languages $K$.
Within a configuration, each target language $t$ receives $D_t$ training tokens.
We seek a function $L_t(N,D_t,E,K)$ that predicts the validation loss of target language $t$.
Averaging $L_t$ over the training languages gives the mean loss of a configuration.

\subsection{Core Joint Law}
\label{sec:core-law}
We build on the standard scaling-law decomposition \citep{hoffmann_2022_compute_optimal},
\begin{equation}
    L(N,D) = L_{\infty} +
    \frac{A}{N^{\alpha}} +
    \frac{B}{D^{\beta}},
    \label{eq:chinchilla}
\end{equation}
which splits loss into an asymptotic floor $L_{\infty}$, reflecting the entropy of natural text \citep{shannon_1948_communication}, plus a model-limited term that decreases with the number of parameters $N$ and a data-limited term that decreases with training tokens $D$.
The coefficients $A,B$ set the magnitude of these terms, while the exponents $\alpha,\beta$ set their scaling rates.

JointMoE \citep{ludziejewski_2025_joint_moe} extends this form with the number of experts, whereas ATLAS \citep{longpre_2026_atlas} adds training tokens per language and language coverage.
Building on both forms, we introduce a flexible \emph{core joint law}, which models $L_t(N,D_t,E,K)$ as
\begin{equation}
    L_{\infty,t}
    + \frac{A_t E^\delta K^{\phi(E)}}{N^{\alpha(E)}}
    + \frac{B_t E^\omega K^{\psi(E)}}{D_t^{\beta(E)}} .
    \label{eq:joint-core}
\end{equation}
The language-specific coefficients $L_{\infty,t}$, $A_t$, and $B_t$ capture differences between target languages.
Following JointMoE, the factors $E^\delta$ and $E^\omega$ scale both loss terms with the number of experts.
Following ATLAS, the factors $K^{\phi(E)}$ and $K^{\psi(E)}$ scale both terms with language coverage.
We additionally allow their exponents to depend on $E$, and test whether this flexibility improves held-out prediction.

Following the log-linear exponents of JointMoE, we parameterize each exponent $q\in\{\alpha,\beta,\phi,\psi\}$ as
\begin{equation}
    q(E) = a_q + b_q \ln(E),
    \label{eq:expert-exponents}
\end{equation}
so each exponent responds to the number of experts through its own intercept $a_q$ and slope $b_q$.
Appendix~\ref{sec:expert-count-conventions} compares this choice with alternative parameterizations of $E$.

\subsection{Predictive Refinements}
\label{sec:refinements}
The core law makes three simplifying assumptions that become limiting in our training regime.
Following prior empirical scaling-law studies \citep{ludziejewski_2024_fine_grained_moe,ludziejewski_2025_joint_moe,longpre_2026_atlas}, we replace each assumption with a more flexible term and evaluate each change on held-out configurations (Section~\ref{sec:result-selection}).

\paragraph{Parameter-data coupling.}
The core law treats the effects of the number of active parameters and training tokens per language as separable.
Early joint laws couple the two \citep{rosenfeld2020a,kaplan_2020_scaling_laws}, and recent work reintroduces this coupling \citep{li2025predictable}.
We add a factor $N^\zeta$ to the data-limited term, allowing additional model capacity to change the benefit of additional training tokens per language.

\paragraph{Language-coverage-dependent parameter scaling.}
Prior work links the curse of multilinguality to limited model capacity \citep{conneau_2020_xlmr,chang_2024_multilinguality_curse}, which suggests that the return on additional active parameters depends on language coverage.
ATLAS models this capacity cost through the factor $K^{\phi}$ but keeps the exponent of $N$ fixed.
We also let this exponent vary with language coverage,
\begin{equation}
    \alpha(E,K) = \alpha(E) + \gamma_N g(K),
    \label{eq:language-coverage-parameter}
\end{equation}
where $g(K)=\max\{0,\log(K/K_0)\}$ grows with log language coverage above a fixed reference coverage $K_0$, matching the power-law coverage factors of the core law.
The coefficient $\gamma_N$ sets how coverage shifts the exponent of $N$, and $\gamma_N=0$ recovers the core law.

\paragraph{Data repetition.}
Long training runs can revisit the same target-language data multiple times.
Following ATLAS, which adapts the effective-data model of \citet{muennighoff_2023_data_constrained}, we replace training tokens per language $D_t$ with an effective exposure $D_t^{\mathrm{eff}}$.
Let $U_t$ denote the unique tokens available for language $t$ and $R_t=\max(D_t/U_t-1,0)$ the number of repeated passes.
We define
\begin{equation}
    D_t^{\mathrm{eff}} = \min(D_t,U_t) + \frac{U_t}{r}\left(1-e^{-rR_t}\right),
    \label{eq:effective-exposure}
\end{equation}
so the first pass counts fully, while subsequent passes contribute progressively less.
The repetition parameter $r=\lambda_0 P^{\rho}$ depends on the total number of stored parameters $P$, which lets models with more stored parameters discount repetition differently. 

\paragraph{Selected law.}
Combining the three refinements while keeping the language coverage exponents $\phi$ and $\psi$ independent of $E$ gives our selected \emph{repetition-aware joint law}, which models $L_t(N,D_t,E,K)$ as
\begin{equation}
    L_{\infty,t}
    + \frac{A_t E^\delta K^{\phi}}{N^{\alpha(E,K)}}
    + \frac{B_t E^\omega K^{\psi} N^\zeta}{(D_t^{\mathrm{eff}})^{\beta(E)}} .
    \label{eq:joint-selected}
\end{equation}
It has 12 shared coefficients plus three coefficients per target language (Appendix~\ref{sec:law-parameterization}).
We also evaluate a composition-aware extension that groups training tokens from other languages by shared language family and script (Appendix~\ref{sec:interaction-ablation}).

\section{Experimental Setup}
\label{sec:experimental-design}
\begin{figure*}[t]
    \centering
    \includegraphics[width=\textwidth]{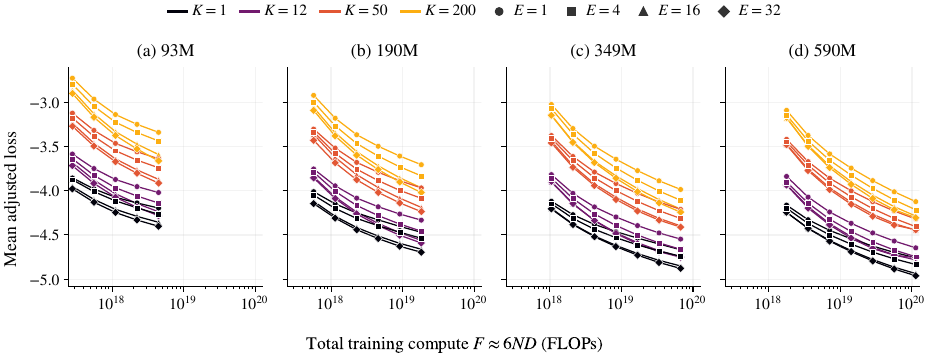}
    \caption{Mean unigram-adjusted validation loss across 64 training configurations and 400 endpoints, averaged over training languages.}
    \label{fig:compute-overview}
\end{figure*}
\subsection{Model Architecture and Training}
All experiments use decoder-only Transformer language models \citep{vaswani_2017_attention} with SwiGLU feed-forward layers \citep{shazeer_2020_glu_variants}.
Sparse configurations replace each feed-forward block with a Top-1 mixture-of-experts layer.
We vary the number of experts over $E\in\{1,4,16,32\}$, with $E=1$ as the dense baseline.
All models share one SentencePiece Unigram tokenizer with a 64k vocabulary \citep{kudo_2018_sentencepiece}, which we train on our multilingual training data sample.
We adopt four depth and width settings from \citet{ludziejewski_2025_joint_moe}, which give $N\in\{93,190,349,590\}$ active parameters (in millions) including embeddings.
Increasing $E$ keeps $N$ approximately fixed while raising the total number of stored parameters. 
Further training details and the learning-rate schedule are provided in Appendix~\ref{sec:appendix-fit}.

%
%
%
\subsection{Data and Sampling}
We train each model on four nested language mixtures with $K\in\{1,12,50,200\}$ languages, so every smaller set is contained in each larger one.
The smallest mixture contains only English, while the 12- and 50-language sets follow ATLAS.
The largest set adds the 150 remaining FineWeb2 languages with the most training data, so coverage grows roughly fourfold at each step.
English data come from FineWeb, while all other languages come from FineWeb2 \citep{penedo_2024_fineweb,penedo_2025_fineweb2}.
Appendix~\ref{sec:appendix-tiers} lists every language.

Within each mixture, we sample languages with temperature smoothing at exponent $0.3$ \citep{conneau_2020_xlmr,xue_2021_mt5}.


%

%
\subsection{Training Grid and Fitting}
Combining the four values each of $N$, $E$, and $K$ results in 64 training configurations.
We call the total number of training tokens $D$ at which a model is evaluated its training horizon.
Larger models train for longer, with maximum training horizons of 8, 16, 32, and 32B tokens for increasing $N$.
Following \citet{ludziejewski_2025_joint_moe}, each run also branches at shorter training horizons that double from 0.5B tokens.
Each branch ends with its own learning-rate cooldown \citep{hagele2024scaling}, and we call the resulting model for that training horizon an endpoint.
Evaluating the 400 endpoints on their respective training languages yields 26,300 per-language loss observations.

As in other compute-intensive scaling-law studies \citep{hoffmann_2022_compute_optimal,ludziejewski_2025_joint_moe,longpre_2026_atlas}, we train each configuration in our main experiments with one seed.
Seed noise therefore enters the law fits as residual error, which the held-out evaluation already measures.
For the interaction between experts and language coverage, we follow the targeted seed checks of \citet{clark_2022_unified_scaling} and replicate our main comparison across additional seeds.
At $N\approx190$M, we retrain $E\in\{1,16,32\}$ crossed with $K\in\{1,12,200\}$ under two additional seeds, which spans dense to sparse models and monolingual to the broadest coverage (Appendix~\ref{par:seed-replication}).

We compare multiple scaling laws on held-out configurations, withholding an entire edge of the training grid to test generalization to configurations excluded from fitting.
The three splits hold out $N\approx590$M, $E=32$, or $K=200$.
We report RMSE with equal weight per configuration, averaged over the three splits (Appendix~\ref{sec:reproduction-specification}).
We additionally evaluate a model with 1.15B active and 21.23B stored parameters as a stress test.
\subsection{Compute and Evaluation}
We measure training compute $F$ in logical FLOPs and approximate it as $F\approx6ND$ \citep{hoffmann_2022_compute_optimal,ludziejewski_2025_joint_moe}.
This excludes routing and communication overheads, which Appendix~\ref{sec:appendix-resource-systems} reports separately.

Following vocabulary-insensitive evaluation in multilingual scaling \citep{tao_2024_vocabulary_scaling,longpre_2026_atlas}, the loss $L_t$ that our laws predict is the unigram-adjusted validation loss
\begin{equation}
    L_t = -\frac{1}{T_t} \sum_{i=1}^{T_t} \log
    \frac{p_\theta(w_i\mid w_{<i})}
    {p_{\mathrm{uni},t}(w_i)},
\end{equation}
where $T_t$ is the number of scored validation tokens, $w_i$ the token at position $i$, $p_\theta(w_i\mid w_{<i})$ its probability under the language model, and $p_{\mathrm{uni},t}(w_i)$ its probability under a fixed unigram model estimated from the training data of language $t$.

We also evaluate 14 multilingual benchmarks that cover language modeling, grammaticality, reasoning, and classification (Appendix~\ref{sec:appendix-benchmarks}).
\section{Results}
\label{sec:results}
We first validate the joint law on held-out configurations, then use measurements and the fitted law to answer the three research questions.
As an overview, Figure~\ref{fig:compute-overview} shows the mean loss across 64 configurations and 400 training endpoints.
These averages combine different languages and numbers of training tokens per language, so the analyses below work with per-language losses instead.

\subsection{Scaling Law Evaluation}
\label{sec:result-selection}

\begin{table*}[t]
\centering\small
\renewcommand{\arraystretch}{1.08}
\setlength{\tabcolsep}{4pt}
\begin{tabular*}{\textwidth}{@{\extracolsep{\fill}}lrrrrr@{}}\toprule
& \multicolumn{3}{c}{Held-out configurations} & & \\
\cmidrule(lr){2-4}
Scaling law & $N\approx590$M & $E=32$ & $K=200$ & Mean RMSE & Larger model \\
\midrule
\addlinespace[2pt]
\rowcolor{black!5}[0pt][0pt]\multicolumn{6}{@{}l@{}}{\strut\hspace{3pt}\textit{Baselines with original inputs}} \\
Chinchilla ($D$) & 0.10630 & 0.13127 & 0.22741 & 0.15499 & 0.23611\\
JointMoE ($D$) & 0.08943 & 0.08668 & 0.20957 & 0.12856 & 0.25269\\
Adaptive ATLAS ($N$) & 0.07088 & 0.11284 & 0.10160 & 0.09511 & 0.23349\\
\addlinespace[2pt]
\rowcolor{black!5}[0pt][0pt]\multicolumn{6}{@{}l@{}}{\strut\hspace{3pt}\textit{Baselines with alternative inputs}} \\
Chinchilla ($D_t$) & 0.07147 & 0.11326 & 0.10171 & 0.09548 & 0.23403\\
JointMoE ($D_t$) & 0.04241 & 0.03342 & 0.06267 & 0.04617 & 0.24733\\
Adaptive ATLAS ($P$) & 0.07257 & 0.09142 & 0.08057 & 0.08152 & 0.25556\\
\addlinespace[2pt]
\rowcolor{black!5}[0pt][0pt]\multicolumn{6}{@{}l@{}}{\strut\hspace{3pt}\textit{Our joint laws}} \\
Core joint law & 0.05859 & 0.03830 & 0.05511 & 0.05067 & 0.27880\\
+ parameter--data coupling + coverage-dependent scaling & 0.03533 & 0.03049 & 0.04507 & 0.03696 & 0.31586\\
\textbf{Repetition-aware joint law (ours)} & \textbf{0.02921} & \textbf{0.02576} & \textbf{0.04461} & \textbf{0.03319} & \textbf{0.17298}\\
\bottomrule\end{tabular*}
\caption{
Prediction RMSE across three held-out settings. The mean column averages the three holdouts. The larger model column reports equal-language RMSE for a separate model with 1.15B active and 21.23B stored parameters.}
\label{tab:joint-story-validation}
\end{table*}

Table~\ref{tab:joint-story-validation} compares our core and refined joint laws with Chinchilla, JointMoE, and ATLAS.
We first evaluate these baselines with their original inputs, then adapt them to our multilingual MoE setting by using training tokens per language $D_t$ instead of total tokens $D$ for Chinchilla and JointMoE, and stored parameters $P$ instead of active parameters $N$ for ATLAS.
All baselines are refitted on our data, with further details in Appendix~\ref{sec:appendix-joint-story}.
All evaluated target languages appear in the fitting data, while the held-out model-size, expert-count, or coverage configurations are excluded from coefficient fitting.
A complementary composition-aware law and additional ablations are reported in Appendix Table~\ref{tab:validation-ablations}.

\paragraph{Joint scaling improves prediction across the grid.}
Adding parameter--data coupling ($N^\zeta$) and coverage-dependent scaling (Equation~\ref{eq:language-coverage-parameter}) to the core joint law reduces mean RMSE from 0.05067 to 0.03696.
The final repetition-aware law further reduces RMSE to 0.03319, outperforming the strongest baseline with original inputs, ATLAS ($N$), by 65\% and the strongest baseline with alternative inputs, JointMoE ($D_t$), by 28\%.

\paragraph{Data repetition becomes important beyond the training grid.}
We next use a larger model with 1.15B active parameters, 21.23B total parameters, and 32 experts, trained on 200 languages for 68.7B tokens, as a stress test beyond the initial training experiments.
On this larger and longer run, the core joint law is less accurate than the evaluated Chinchilla and ATLAS baselines.
The largest errors occur for languages whose exposure substantially exceeds their available unique data.
Accounting for repetition reduces RMSE from 0.31586 to 0.17298 and outperforms both baselines.
This highlights that extrapolation to longer training requires distinguishing total training tokens from unique training data.
Our repetition-aware joint law also transfers to changed language mixtures, reducing prediction RMSE by 74\% relative to the strongest evaluated baseline (Appendix~\ref{sec:composition-study}).
\subsection{Research Questions}
\label{sec:result-rqs}
\begin{figure*}[t]
    \centering
    \includegraphics[width=\textwidth]{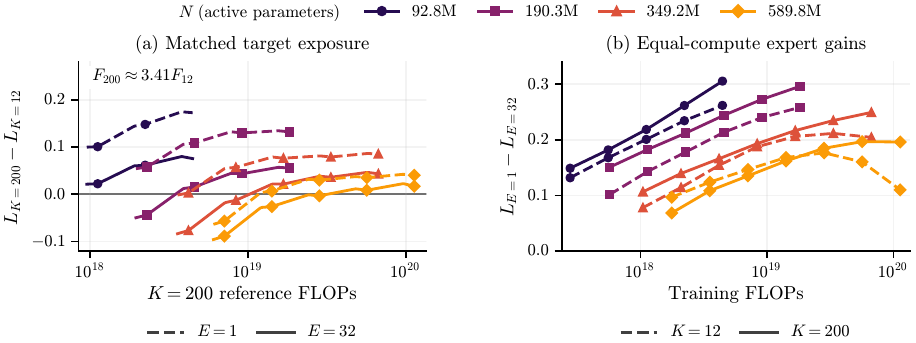}
    \caption{Mean loss differences for the same 12 languages in models trained on 12 or 200 languages.
    (a) Difference in loss at matched training tokens per language. Negative values favor broader language coverage.
    (b) Loss reduction from increasing the number of experts from 1 to 32 at equal training compute.}
    \label{fig:joint-capacity-response}
\end{figure*}
We next examine the implications of broader language coverage for model design and language quality, covering each research question one by one. 

\subsubsection{Language Coverage Cost (RQ1)}
\label{sec:result-pressure}


We measure the change in loss for the same 12 languages when expanding training from 12 to 200 languages, at matched $N$, $E$, and training tokens per language ($D_t$), following the approach of \citet{pfeiffer_2022_lifting_curse}.
Because language sampling differs across mixtures, the number of training tokens from other languages can differ.
We therefore interpret the result as an effect of training on broader mixtures.
We quantify this change in loss as
$\Delta L_t = L_t(K=200)-L_t(K=12)$, where positive values indicate a language coverage penalty, while negative values indicate a benefit from broader language coverage.


Figure~\ref{fig:joint-capacity-response} summarizes this interaction over training.
Panel (a) shows the difference in loss at matched training tokens per language for models with 1 and 32 experts at each active model size.
Panel (b) shows the loss reduction from increasing the number of experts from 1 to 32 at equal training compute.
The loss increase of expanding from 12 to 200 languages becomes substantially smaller as the number of experts increases.
Across the 12 languages shared between $K=12$ and $K=200$, the mean penalty falls from $0.0796$ at $E=1$ to $0.0648$ at $E=4$, $0.0321$ at $E=16$, and $0.0135$ at $E=32$.
Increasing the number of experts from 1 to 32 therefore reduces the loss penalty of expanding from 12 to 200 languages by 83\% (Appendix Figure~\ref{fig:pressure}).

We interpolate the $K=12$ reference in $\log D_t$, average comparisons at each value of $N$, and weight the four values equally.
To test sensitivity to mixture composition, we repeat the comparison with alternative 12-language mixtures and find that additional experts reduce the loss increase by 76.6 to 93.3\% (Appendix~\ref{sec:composition-coverage}).

At 589.8M active parameters, the $K=200$ training initially gives a lower loss than $K=12$ for both $E=1$ and $E=32$, but this benefit becomes a penalty as training continues (Figure~\ref{fig:joint-capacity-response}a).
Increasing from $E=1$ to $E=32$ delays this crossover by approximately $2.7\times$ compute.
The interaction varies with training duration and what is held fixed (Figure~\ref{fig:joint-capacity-response}b, Appendix~\ref{sec:capacity-duration-checks}).



At approximately 190M active parameters, increasing the number of experts from 1 to 32 reduces the loss penalty of expanding from 12 to 200 languages by 83.6\%, 82.6\%, and 66.4\% across three training seeds (Appendix Figure~\ref{fig:pressure-seeds}).
The reduction in loss increase remains consistent across interpolation methods, learning-rate controls, and other language-coverage expansions.


\paragraph{External evaluation supports the loss interaction.}
\label{sec:result-benchmark-transfer}
On the same 12 trained languages at $K=12$ and $K=200$, increasing the number of experts from 1 to 32 reduces the increase in external text loss by 22 to 57\% on FLORES and 20 to 62\% on NTREX across all four settings of $N$.
The direction is consistent across all three 190M seeds and under byte normalization.

The interaction also appears in downstream accuracy for MultiBLiMP, where expert gains are larger at $K=200$.
For HellaSwag and XStoryCloze, expert expansion improves accuracy at both coverage levels, but gains are smaller at $K=200$.
Thus, the loss interaction extends beyond the validation corpus, while its effect in downstream accuracy depends on the task (Appendix~\ref{sec:appendix-benchmarks}).

\subsubsection{Capacity Allocation (RQ2)}
\label{sec:result-allocation}
\begin{figure*}[t]
    \centering
    \includegraphics[width=\textwidth]{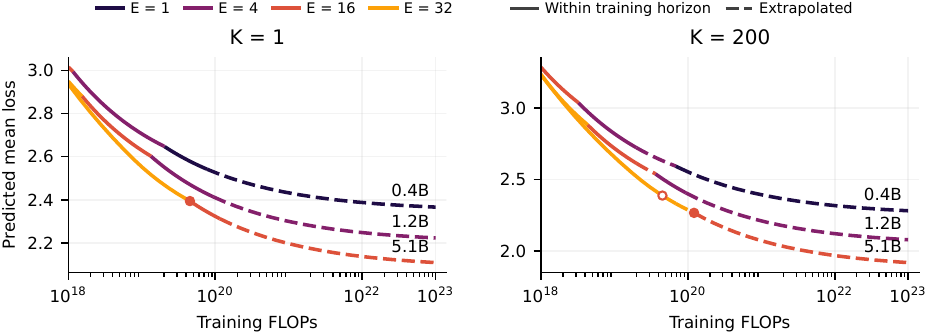}
    \caption{Minimum predicted mean loss from the composition-aware scaling-law under different total-parameter budgets, with 92.8 to 589.8M active parameters. Loss is averaged equally across its training languages.
    Filled circles indicate the $E=32\to16$ switch at the 5.1B-parameter budget.}
    \label{fig:fitted-allocation-language-shift}
\end{figure*}

Figure~\ref{fig:joint-capacity-response}a shows that either increasing the number of active parameters or the number of experts reduces the loss increase from broader language coverage.
However, additional active parameters increase training compute, while additional experts primarily increase stored parameters.
Previous work shows that the preferred balance between the number of active parameters and the number of experts depends on training-compute and total-parameter budgets \citep{ludziejewski_2025_joint_moe}. 
At each compute budget, Figure~\ref{fig:fitted-allocation-language-shift} uses the composition-aware law to select the lowest predicted mean loss under a total-parameter budget.

\paragraph{Broader language coverage favors expert-heavy allocations for longer.}
Under a 0.4B total-parameter limit, the preferred measured configuration switches from 92.8M active parameters with 16 experts to 190.3M with 4 experts.
The switch occurs at $1.70\times10^{18}$ FLOPs for 12 languages and $2.75\times10^{18}$ FLOPs for 200 languages, a $1.62\times$ delay.
These estimates use smoothed curves on the same 12 languages and remain within the observed training range (Appendix~\ref{sec:allocation-details}).

The fitted allocation curves show the same pattern over a wider range of storage budgets.
At a 5.1B total-parameter budget, the predicted $E=32\rightarrow16$ switch occurs at $4.51\times10^{19}$ FLOPs for $K=1$ and $1.22\times10^{20}$ FLOPs for $K=200$, a $2.70\times$ delay, with the latter predicted beyond the observed training horizon.


\subsubsection{Per-Language Preservation (RQ3)}
\label{sec:result-preservation}
Starting from a reference model trained on 50 languages with $2\times10^{18}$ FLOPs and a 2.2B total-parameter limit, expanding to 200 languages recovers the original 50-language mean loss at $1.92\times$ the reference compute.
Both configurations use approximately 190.3M active parameters and 32 experts.

At the mean-preserving point, only 36\% of the original languages match or improve upon their 50-language reference, leaving 32 languages with higher loss.
A supported configuration matches or improves the reference loss for all 50 languages at $2.32\times$ compute (Appendix Figure~\ref{fig:allocation-preservation-detail}), although this need not be the minimum required budget.


In this comparison, matching training tokens per language, preserving mean quality, and preserving all 50 languages require approximately $1.75\times$, $1.92\times$, and $2.32\times$ compute, respectively.
Additional budgets and language-expansion settings are reported in Appendix~\ref{sec:allocation-details}.




\section{Discussion}
\label{sec:discussion}
Our results suggest that language coverage should be treated as a capacity-allocation variable in multilingual MoEs, with broader coverage increasing the relative value of expert capacity under the resource budgets we study.


Matching training tokens per language, preserving average multilingual quality, and preserving the quality of every original language require different amounts of compute.
Recovering the average can still leave many languages worse off, so multilingual capacity planning should therefore specify which notion of preservation is required rather than treating aggregate quality as sufficient.

The predictive refinements highlight that scaling laws should be evaluated for the decisions they support.
Accounting for repeated training data improves extrapolation to longer training runs, while accounting for language composition improves predictions when the language mixture changes.

Our allocation analysis uses logical compute $F\approx6ND$, following standard scaling-law accounting \citep{kaplan_2020_scaling_laws,hoffmann_2022_compute_optimal} and abstracting from hardware-specific memory, communication, and parallelism costs, which are reported separately in Appendix~\ref{sec:appendix-resource-systems}.


\section{Conclusion}
\label{sec:conclusion}

We study how the number of active parameters, training tokens per language, the number of experts, and language coverage interact in massively multilingual MoE pretraining.
Our central finding is that expert expansion changes the effect of increased language coverage rather than merely lowering loss overall.
At matched training tokens per language, increasing from $E=1$ to $E=32$ reduces the loss penalty of expanding from 12 to 200 languages by 83\%, with the same direction observed across additional seeds, nested coverage transitions, and external text evaluation.

We introduce a joint scaling law over $N$, $D_t$, $E$, and $K$ that improves loss prediction on held-out configurations.
Coverage-aware and repetition-aware terms improve held-out prediction, including on a larger-model stress test.
Preserving average multilingual quality, however, can still leave many individual languages worse off.

Together, our results show that broader language coverage changes both how much capacity is needed and how that capacity should be allocated.
Multilingual MoE pretraining is therefore best viewed as a joint allocation problem over active capacity, expert capacity, training tokens per language, and language coverage.

\ns{Limitations and Ethical Considerations do not count for 8-page limit: \url{https://aclrollingreview.org/authorchecklist\#paper-format} }
\section{Limitations}
\label{sec:limitations}
As is common in compute-intensive scaling-law studies \citep{hoffmann_2022_compute_optimal,ludziejewski_2025_joint_moe,longpre_2026_atlas}, the main factorial grid uses one training seed.
We therefore repeat the main comparison at matched training tokens with two additional seeds at selected $E$ and $K$ settings.
The main effect remains consistent in direction across all three seeds.

Our conclusions apply to the studied decoder-only, Top-1 MoE setting with fixed expert granularity, tokenizer, corpus, sampling scheme, and $E\leq32$.
Increasing the number of experts also changes stored capacity and the expert-aware training recipe.

We assess the scaling law on held-out configurations across the number of active parameters, the number of experts, and language coverage (Appendix~\ref{sec:reproduction-specification}).
Its transfer to substantially different architectures, corpora, or training regimes remains outside the scope of this study.
Finally, our allocation results use logical compute and total parameters. 
Actual hardware cost additionally depends on communication, memory, and parallelism overheads. 
\section*{Ethical Considerations}
\label{sec:ethics}

Our results show that preserving average multilingual quality does not necessarily preserve every individual language.
Average performance should therefore not be interpreted as equal quality or representation across languages or communities.

The experiments use existing web-derived corpora and may inherit their biases and other data-quality limitations.
We do not evaluate the trained checkpoints for deployment safety, privacy, or bias.
Sparse models may also incur hardware and energy costs beyond logical FLOPs.
No new human participants or annotations were used.
%
%
%
%


\bibliography{bib/custom,bib/anthology-paper}
\appendix
\clearpage
\jd{work in progress. main paper mostly finished from my side, still refining the appendix}
\jd{TODO: wording consistency checks: endpoints / configuraitons / trejctories, strucurred holdouts / validaiton splits / hoold our configuraitons, matched exposure etc.}
\section*{Appendix contents}
\begingroup
\raggedright
\newcommand{\appendixentry}[1]{%
    \par\noindent\hangindent=1.5em
    \hyperref[#1]{\ref*{#1}\quad\nameref*{#1}}\nobreak\dotfill\nobreak\pageref{#1}\par}
\appendixentry{sec:appendix-scope}
\appendixentry{sec:appendix-selection}
\appendixentry{sec:appendix-seeds}
\appendixentry{sec:composition-study}
\appendixentry{sec:appendix-allocation}
\appendixentry{sec:appendix-benchmarks}
\appendixentry{sec:appendix-routing}
\appendixentry{sec:appendix-resource-systems}
\endgroup

\section{Experimental Details}
\label{sec:appendix-scope}
Our main experiment varies four values for the number of active parameters ($N\in\{92.8,190.3,349.2,589.8\}\,\mathrm{M}$), four numbers of experts ($E\in\{1,4,16,32\}$), and four language-coverage settings ($K\in\{1,12,50,200\}$), yielding 64 model configurations. Each configuration is trained for 8--32B tokens, depending on the number of active parameters, with a separate cooldown at each of five to seven training horizons, resulting in 400 endpoints in total.

We additionally train 18 runs with additional seeds, 18 runs on alternative $K=12$ language mixtures, and one larger-model stress test.
Together, the core study therefore contains 101 training runs and 617 evaluated checkpoints.

Separate learning-rate experiments verify that the expert-aware learning-rate rule used in the main study is appropriate across the evaluated expert and language-coverage settings.
Table~\ref{tab:experiment-inventory} summarizes the training experiments.

\begin{table}[!htbp]
    \centering\small
    \begin{tabular}{@{}lrr@{}}
    \toprule
    Study component & Runs & Endpoints \\
    \midrule
    \multicolumn{3}{@{}l}{\textit{Core study and extensions}} \\
    Primary factorial grid & 64 & 400 \\
    Additional training seeds & 18 & 108 \\
    Alternative $K=12$ mixtures & 18 & 108 \\
    Larger-model stress test & 1 & 1 \\
    \textbf{Subtotal} & \textbf{101} & \textbf{617} \\
    \midrule
    \multicolumn{3}{@{}l}{\textit{Learning-rate validation and controls}} \\
    Learning-rate validation experiments & 108 & 108 \\
    Fixed-learning-rate control grid & 64 & 400 \\ 
    \bottomrule
    \end{tabular}
    \caption{Training runs and evaluated endpoints across the study.}
    \label{tab:experiment-inventory}
\end{table}

\paragraph{Seed replication.}
\label{par:seed-replication}
Following the targeted seed checks of \citet{clark_2022_unified_scaling}, we repeat nine configurations at 190.3M active parameters with $E\in\{1,16,32\}$ and $K\in\{1,12,200\}$.
Using three training seeds yields 18 additional runs and 108 evaluated checkpoints across six training horizons from 0.5B to 16B tokens.

\subsection{Data and language sets}
\label{sec:appendix-tiers}

We use four nested language sets, with $K\in\{1,12,50,200\}$ and each smaller set contained in the larger ones.
This allows us to compare the same languages under increasingly broader multilingual mixtures.
Table~\ref{tab:language-membership} lists all language--script pairs.

The $K=1$ setting contains English only, while the $K=12$ and $K=50$ sets follow ATLAS.
We extend the $K=50$ set to $K=200$ with 150 additional FineWeb2 languages selected based on data availability, following the data-driven language coverage used in large multilingual pretraining such as XLM-R and mT5 \citep{conneau_2020_xlmr,xue_2021_mt5}.
We additionally use two alternative $K=12$ language sets to assess sensitivity to language composition. Their construction and evaluation are described in Appendix~\ref{sec:composition-study}.

Following the smoothed language sampling of XLM \citep{conneau_2019_crosslingual} and XLM-R \citep{conneau_2020_xlmr}, we define the sampling probability of language $t$ as
\begin{equation}
 p_t(K)=\frac{n_t^{0.3}}{\sum_{\ell\in\mathcal{L}_K}n_\ell^{0.3}},
 \label{eq:sampling}
\end{equation}
where $\mathcal{L}_K$ is the selected language set and $n_t$ is the number of available training tokens of language $t$.
The exponent increases the sampling probability of low-resource languages and thus reduces the bias towards high-resource languages.
English has a sampling probability of 100\%, 19.38\%, 9.94\%, and 5.68\% at $K=1,12,50,200$, respectively.

Following ATLAS \cite{longpre_2026_atlas}, we construct a document-level validation split for each language containing approximately 20 million tokens or 20\% of the available unique tokens for that language, whichever is smaller.
Training and validation sets are disjoint at the document level.

We use a shared 64,000-piece SentencePiece Unigram tokenizer for all training experiments.
It is trained once on 100 million sentences across all 200 languages, using the same language sampling exponent of 0.3.
Byte fallback is enabled.
To make validation loss more comparable across languages, we subtract a language-specific unigram baseline that accounts for differences in token frequencies. The baseline is estimated from the training data only. Special tokens and padding are excluded from the loss.

\begin{table*}[t]
\centering\small
\begin{tabular}{@{}p{\textwidth}@{}}
\toprule
\raggedright \textbf{$K=12$ (12):} \mbox{cmn\_Hani}, \mbox{deu\_Latn}, \mbox{eng\_Latn}, \mbox{fra\_Latn}, \mbox{hin\_Deva}, \mbox{ita\_Latn}, \mbox{jpn\_Jpan}, \mbox{pol\_Latn}, \mbox{por\_Latn}, \mbox{rus\_Cyrl}, \mbox{spa\_Latn}, \mbox{vie\_Latn}\tabularnewline
\addlinespace[2pt]
\raggedright \textbf{Added at $K=50$ (38):} \mbox{afr\_Latn}, \mbox{arb\_Arab}, \mbox{bak\_Cyrl}, \mbox{bam\_Latn}, \mbox{ben\_Beng}, \mbox{cat\_Latn}, \mbox{ckb\_Arab}, \mbox{cym\_Latn}, \mbox{ell\_Grek}, \mbox{ewe\_Latn}, \mbox{fas\_Arab}, \mbox{fil\_Latn}, \mbox{gaz\_Latn}, \mbox{gug\_Latn}, \mbox{guj\_Gujr}, \mbox{hau\_Latn}, \mbox{heb\_Hebr}, \mbox{hrv\_Latn}, \mbox{ind\_Latn}, \mbox{jav\_Latn}, \mbox{kaz\_Cyrl}, \mbox{kor\_Hang}, \mbox{mar\_Deva}, \mbox{mya\_Mymr}, \mbox{nld\_Latn}, \mbox{nob\_Latn}, \mbox{pbt\_Arab}, \mbox{ron\_Latn}, \mbox{smo\_Latn}, \mbox{swe\_Latn}, \mbox{swh\_Latn}, \mbox{tel\_Telu}, \mbox{tha\_Thai}, \mbox{tur\_Latn}, \mbox{ukr\_Cyrl}, \mbox{urd\_Arab}, \mbox{yor\_Latn}, \mbox{zsm\_Latn}\tabularnewline
\addlinespace[2pt]
\raggedright \textbf{Added at $K=200$ (150):} \mbox{aeb\_Arab}, \mbox{als\_Latn}, \mbox{amh\_Ethi}, \mbox{anp\_Deva}, \mbox{apc\_Arab}, \mbox{arg\_Latn}, \mbox{ars\_Arab}, \mbox{ary\_Arab}, \mbox{arz\_Arab}, \mbox{asm\_Beng}, \mbox{ast\_Latn}, \mbox{azb\_Arab}, \mbox{azj\_Latn}, \mbox{ban\_Latn}, \mbox{bar\_Latn}, \mbox{bel\_Cyrl}, \mbox{bew\_Latn}, \mbox{bho\_Deva}, \mbox{bod\_Tibt}, \mbox{bos\_Latn}, \mbox{bre\_Latn}, \mbox{bul\_Cyrl}, \mbox{bxr\_Cyrl}, \mbox{ceb\_Latn}, \mbox{ces\_Latn}, \mbox{cfm\_Latn}, \mbox{che\_Cyrl}, \mbox{chv\_Cyrl}, \mbox{cnh\_Latn}, \mbox{cos\_Latn}, \mbox{crh\_Latn}, \mbox{dan\_Latn}, \mbox{div\_Thaa}, \mbox{dzo\_Tibt}, \mbox{ekk\_Latn}, \mbox{epo\_Latn}, \mbox{eus\_Latn}, \mbox{fao\_Latn}, \mbox{fij\_Latn}, \mbox{fin\_Latn}, \mbox{fry\_Latn}, \mbox{fur\_Latn}, \mbox{gla\_Latn}, \mbox{gle\_Latn}, \mbox{glg\_Latn}, \mbox{glk\_Arab}, \mbox{gmh\_Latn}, \mbox{grc\_Grek}, \mbox{gsw\_Latn}, \mbox{hac\_Arab}, \mbox{hat\_Latn}, \mbox{haw\_Latn}, \mbox{hbo\_Hebr}, \mbox{hif\_Latn}, \mbox{hil\_Latn}, \mbox{hne\_Deva}, \mbox{hsb\_Latn}, \mbox{hun\_Latn}, \mbox{hye\_Armn}, \mbox{hyw\_Armn}, \mbox{iba\_Latn}, \mbox{ibo\_Latn}, \mbox{ido\_Latn}, \mbox{ilo\_Latn}, \mbox{isl\_Latn}, \mbox{kaa\_Cyrl}, \mbox{kab\_Latn}, \mbox{kal\_Latn}, \mbox{kan\_Knda}, \mbox{kat\_Geor}, \mbox{kha\_Latn}, \mbox{khk\_Cyrl}, \mbox{khm\_Khmr}, \mbox{kin\_Latn}, \mbox{kir\_Cyrl}, \mbox{kiu\_Latn}, \mbox{kmr\_Latn}, \mbox{lao\_Laoo}, \mbox{lat\_Latn}, \mbox{lim\_Latn}, \mbox{lin\_Latn}, \mbox{lit\_Latn}, \mbox{lmo\_Latn}, \mbox{ltg\_Latn}, \mbox{ltz\_Latn}, \mbox{lug\_Latn}, \mbox{lus\_Latn}, \mbox{lvs\_Latn}, \mbox{mai\_Deva}, \mbox{mal\_Mlym}, \mbox{mhr\_Cyrl}, \mbox{mkd\_Cyrl}, \mbox{mlt\_Latn}, \mbox{mri\_Latn}, \mbox{mww\_Latn}, \mbox{nap\_Latn}, \mbox{nde\_Latn}, \mbox{nds\_Latn}, \mbox{new\_Deva}, \mbox{nno\_Latn}, \mbox{npi\_Deva}, \mbox{nrm\_Latn}, \mbox{nya\_Latn}, \mbox{oci\_Latn}, \mbox{ory\_Orya}, \mbox{oss\_Cyrl}, \mbox{pan\_Guru}, \mbox{pap\_Latn}, \mbox{plt\_Latn}, \mbox{pms\_Latn}, \mbox{pnb\_Arab}, \mbox{rmn\_Latn}, \mbox{rmy\_Latn}, \mbox{roh\_Latn}, \mbox{rue\_Cyrl}, \mbox{run\_Latn}, \mbox{sah\_Cyrl}, \mbox{san\_Deva}, \mbox{scn\_Latn}, \mbox{sco\_Latn}, \mbox{sdh\_Arab}, \mbox{shn\_Mymr}, \mbox{sin\_Sinh}, \mbox{slk\_Latn}, \mbox{slv\_Latn}, \mbox{sme\_Latn}, \mbox{sna\_Latn}, \mbox{snd\_Arab}, \mbox{som\_Latn}, \mbox{sot\_Latn}, \mbox{srd\_Latn}, \mbox{srp\_Cyrl}, \mbox{sun\_Latn}, \mbox{tam\_Taml}, \mbox{tat\_Cyrl}, \mbox{tgk\_Cyrl}, \mbox{tir\_Ethi}, \mbox{tuk\_Latn}, \mbox{tyv\_Cyrl}, \mbox{udm\_Cyrl}, \mbox{uig\_Arab}, \mbox{uzn\_Latn}, \mbox{uzs\_Arab}, \mbox{wln\_Latn}, \mbox{wuu\_Hani}, \mbox{xho\_Latn}, \mbox{ydd\_Hebr}, \mbox{yue\_Hani}, \mbox{zea\_Latn}, \mbox{zul\_Latn}\tabularnewline
\addlinespace[2pt]
\raggedright \textbf{Compact $K=12$ (12):} \mbox{cmn\_Hani}, \mbox{dan\_Latn}, \mbox{deu\_Latn}, \mbox{eng\_Latn}, \mbox{fra\_Latn}, \mbox{ita\_Latn}, \mbox{nld\_Latn}, \mbox{por\_Latn}, \mbox{ron\_Latn}, \mbox{rus\_Cyrl}, \mbox{spa\_Latn}, \mbox{swe\_Latn}\tabularnewline
\addlinespace[2pt]
\raggedright \textbf{Diverse $K=12$ (12):} \mbox{arb\_Arab}, \mbox{cmn\_Hani}, \mbox{deu\_Latn}, \mbox{eng\_Latn}, \mbox{fra\_Latn}, \mbox{hin\_Deva}, \mbox{jpn\_Jpan}, \mbox{kor\_Hang}, \mbox{por\_Latn}, \mbox{rus\_Cyrl}, \mbox{spa\_Latn}, \mbox{tha\_Thai}\tabularnewline
\bottomrule
\end{tabular}
\caption{Languages included in each training mixture. }
\label{tab:language-membership}
\end{table*}

\paragraph{Licensing and intended use.}
Our work uses FineWeb and FineWeb2 under their respective licenses, along with existing evaluation benchmarks and software.
We do not redistribute the original datasets.
We release configurations, run records, per-language scores, and analysis code under CC BY 4.0 for the results and MIT for the code.
The licenses of the underlying artifacts continue to apply.

The experimental data are provided in the supplementary data package. 
Our training and analysis code is open-source and available online \url{https://anonymous.4open.science/r/multi-moe-scaling-laws/}.
\subsection{Architecture and training}
\label{sec:appendix-fit}

Table~\ref{tab:architecture} gives the four values of $N$. Every feed-forward block uses dropless Top-1 routing when $E>1$, where $E=1$ serves as the dense control.
All models use SwiGLU, RoPE \citep{su_2024_roformer}, RMSNorm \citep{zhang_2019_rmsnorm}, no always-active shared experts, FP32 router probabilities, a load-balancing coefficient of 0.01, a router-$z$ coefficient of 0.001, and no capacity-factor padding.

\begin{table*}[!t]
\centering\small
\begin{tabular}{rrrrrrr}
\toprule
Layers & $d_{\mathrm{model}}$ & Attn. Heads &  $d_{\mathrm{FFN}}$ & $N$ & $N_{\mathrm{nonemb}}$ & Max. tokens \\
\midrule
8  & 512  & 8  & 1536 & 92.8M  & 27.3M  & 8B \\
12 & 768  & 12 & 2304 & 190.3M & 92.0M  & 16B \\
16 & 1024 & 16 & 3072 & 349.2M & 218.1M & 32B \\
20 & 1280 & 20 & 3840 & 589.8M & 426.0M & 32B \\
\bottomrule
\end{tabular}
\caption{Model architecture configurations. $N_{\mathrm{nonemb}}$ indicates active, non-embedding parameters}
\label{tab:architecture}\label{tab:experiments-listing}
\end{table*}

\begin{table*}[!t]
\centering\small
\begin{tabular}{rrrrr}
\toprule
$N$ & $P$, $E=1$ & $E=4$ & $E=16$ & $E=32$ \\
\midrule
92,798,976 & 92,798,976 & 149,422,080 & 375,914,496 & 677,904,384 \\
190,316,544 & 190,316,544 & 381,419,520 & 1,145,831,424 & 2,165,047,296 \\
349,175,808 & 349,175,808 & 802,160,640 & 2,614,099,968 & 5,030,019,072 \\
589,824,000 & 589,824,000 & 1,474,560,000 & 5,013,504,000 & 9,732,096,000 \\
\bottomrule
\end{tabular}
\caption{Exact active ($N$) and total ($P$) parameter counts for each architecture and number of experts in the main experiments, including embeddings.}
\label{tab:stored-parameters}
\end{table*}

Training uses Adam with $\beta_1=0.9$, $\beta_2=0.95$, $\epsilon=10^{-8}$, weight decay 0.1, gradient clipping 1.0, BF16 parameters, and FP32 optimizer states. 
Sequence length is 2,048. Global batches contain 32 sequences below 0.5B tokens, 64 from 0.5B to 1B, and 128 thereafter.
Warmup lasts 130 million tokens, and the primary grid uses seed 1234.
Following \citet{ludziejewski_2025_joint_moe}, for each target duration $D$, training follows a shared trajectory through $0.8D$ tokens and then branches into a continuation that linearly decays the learning rate to zero over the final $0.2D$.
Measurements from the same configuration therefore share training history and are not independent training runs.
Completed-step token counts are used to compute realized training tokens per language and logical training compute.
Training uses one, two, or four NVIDIA H100 GPUs depending on model memory requirements.

\paragraph{Learning-rate choice.}
\begin{figure}[!t]
    \centering
    \includegraphics[width=\columnwidth]{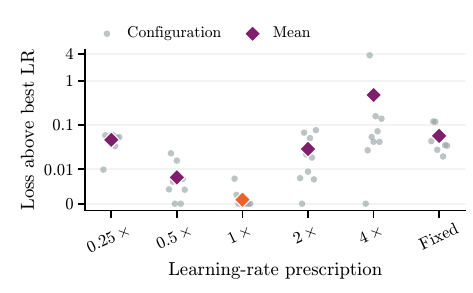}
    \caption{Loss difference relative to the best tested learning rate rule for each configuration.}
    \label{fig:lr-choice}
\end{figure}

To assess the suitability of the learning-rate prescription used in our main experiments, we compare multipliers $\{0.25,0.5,1,2,4\}$ of the expert-aware rule from \citet{ludziejewski_2025_joint_moe}, following the multiplier grid of \citet{bjorck_2025_scaling}.
At 190M active parameters and 2B tokens, we evaluate these rates alongside a fixed rate of $2\times10^{-4}$ across $E\in\{1,16,32\}$ and $K\in\{1,12,200\}$.
The original prescription (multiplier 1) achieves the lowest mean loss among the tested alternatives and outperforms the fixed rate in all nine configurations, supporting its use in the main experiments (Figure~\ref{fig:lr-choice}).

\section{Scaling-Law Specification and Validation}
\label{sec:appendix-selection}
\label{sec:appendix-joint-story}
This section provides the complete specification and validation of our joint scaling law.
We detail the final repetition-aware parameterization and fitting procedure, followed by ablations, sensitivity analysis, structured holdout evaluations, and the larger-model stress test.

\subsection{Final law and fitting procedure}
\label{sec:law-parameterization}
\label{sec:reproduction-specification}
\label{sec:selected-law-specification}

We first restate the repetition-aware joint law and define all quantities used in fitting.
Here, $N$ and $P$ denote active and total stored parameters, respectively.
For numerical fitting, parameter counts (including $P$ in the repetition discount) are expressed in units of $10^8$ and token counts in units of $10^9$.
We also normalize the language-coverage term as $g_{\mathrm{fit}}(K)=\max\{0,\log(K/12)\}/\log(200/12)$.

\begin{equation}
\begin{aligned}
    L_t &= L_{\infty,t}
 + \frac{A_t E^\delta K^{\phi}}
        {N^{\alpha(E,K)}} \\
&\quad + \frac{B_t E^\omega K^{\psi} N^\zeta}
        {(D_t^{\mathrm{eff}})^{\beta(E)}} .
\end{aligned}
\end{equation}

The expert-dependent exponents follow Equation~\ref{eq:expert-exponents}, while the language-coverage adjustment $\alpha(E,K)$ is defined in Section~\ref{sec:scaling-law}.

Following the effective-data model for repeated training data of \citet{muennighoff_2023_data_constrained} and its repetition-aware adaptation in ATLAS \citep{longpre_2026_atlas}, we model repeated passes over the available target-language data with an exponential saturation function.
Let $U_t$ denote the unique training-tokens available for language $t$ before replication. We define
\begin{align}
    R_t&=\max(D_t/U_t-1,0),\\
    S_t&=\min(D_t,U_t)+\frac{U_t}{r}(1-e^{-rR_t}),\\
    r&=\lambda_0(P)^\rho
    \label{eq:selected-effective-exposure}
\end{align}
Here, $R_t$ indicates repeated passes beyond the first pass over the unique available data of a language $U_t$, $S_t$ is the resulting effective number of training tokens per language with repetition parameter $r$ controlling how rapidly additional repeated passes are discounted.
We set $D_t^{\mathrm{eff}}=S_t$ in Equation~\ref{eq:joint-selected} and express effective training tokens per language in billions of tokens, $s_t=S_t/10^9$, for numerical fitting.
The first pass over the available data therefore counts fully, while subsequent passes are progressively discounted.
$P$ is the recorded stored parameter count (Table~\ref{tab:stored-parameters}), determined by the number of active parameters and the number of experts per training configuration. 
The selected law has 12 shared coefficients: two each for $\alpha(E)$ and $\beta(E)$, together with $\delta$, $\omega$, $\phi$, $\psi$, $\gamma_N$, $\zeta$, $\lambda_0$, and $\rho$.
Each target language additionally has three shared coefficients $L_{\infty,t}$, $A_t$, and $B_t$.

Each primary validation split withholds one complete axis value: $N \approx 590$M, $E=32$, or $K=200$.
The remaining 48 configurations are used for fitting and the 16 withheld configurations for evaluation.
For the $K=200$ split, we evaluate only the 50 languages that appear in the fitting data, because the model requires fitted coefficients for each target language.

We fit the scaling laws using configuration-balanced Huber loss \cite{huber_1964_robust}, weighting observations so that each training configuration contributes equally.
For each fitting split, we first fit a common reference scaling law using weighted least squares.
We then set the Huber threshold from the weighted median absolute deviation of its residuals:
\[ h = 1.345 \times 1.4826 \times \mathrm{MAD}_w. \]
Here, $\mathrm{MAD}_w$ measures the median absolute residual deviation around the weighted median.
The factor $1.4826$ scales the MAD to a Gaussian-consistent estimate of residual scale, while $1.345$ is the standard Huber tuning constant, corresponding to approximately 95\% efficiency under Gaussian errors.
This allows the threshold to adapt to the residual scale of each fitting split while using the same calibration rule throughout.
The threshold is computed using fitting observations only and is then held fixed across all candidate scaling-law variants within that split. 
Table~\ref{tab:fit-settings} gives the observation counts and resulting Huber thresholds.

\begin{table}[!t]
    \centering\small
        \begin{tabular}{@{}lrrr@{}}
            \toprule
            Holdout &
            \shortstack{\# Fit\\observations} &
            \shortstack{\# Eval.\\observations} &
            \shortstack{Huber\\threshold $h$} \\
            \midrule
            $N \approx 590$M & 18,936 & 7,364 & 0.1384 \\
            $E=32$               & 19,725 & 6,575 & 0.1168 \\
            $K=200$              & 6,300  & 5,000 & 0.1241 \\
            \bottomrule
        \end{tabular}
        \caption{Numbers of fitting and evaluation observations and Huber thresholds for the three hold-out configurations.}
    \label{tab:fit-settings}
\end{table}

All primary variants use the same observations and weights.
We choose weights so that each configuration contributes equally.
If configuration $j$ contains $m_j$ language/endpoint observations among $T$ fitting configurations, each observation receives weight $w_i=1/(Tm_j)$.
Prediction error is measured using configuration-balanced RMSE, and the three holdout errors averaged equally.

We fit the final repetition-aware law by minimizing
\begin{equation}
    Q=\sum_i w_i\rho_h(\widehat L_i-L_i),
    \label{eq:fitting-objective}
\end{equation}
Here, $Q$ is the fitting objective, $L_i$ and $\widehat L_i$ denote the observed and predicted loss for observation $i$, respectively, and $\rho_h$ denotes the Huber loss with threshold $h$.

We optimize the fitting objective with multistart L-BFGS-B following Chinchilla and JointMoE \citep{hoffmann_2022_compute_optimal,ludziejewski_2025_joint_moe}, calibrating Huber thresholds on each fitting split. 
We optimize the language-specific coefficients for each setting of the shared coefficients. 
Among converged restarts, we select the solution with the lowest fitting objective.
We constrain $A_t$ and $B_t$ to be positive and bound $L_{\infty,t}$ between the unigram-adjusted lower bound and the smallest fitting loss for target language $t$.
Reproduction code provides all parameter bounds, initializations, optimizer settings, and fitted coefficients.

Following JointMoE and ATLAS, we use empirical scaling diagnostics and structured validation to evaluate different scaling-law formulations \citep{ludziejewski_2025_joint_moe,longpre_2026_atlas}.
Every configuration holdout excludes its evaluation configurations from coefficient fitting.
The larger-model run is also excluded from coefficient fitting.

\subsection{Ablations and published-law comparisons}
\label{sec:appendix-prior-baseline}
Table~\ref{tab:validation-ablations} shows contributions of expert-dependent scaling, size--data coupling, language coverage, and language composition.
Each variant is refitted on the same structured validation splits using the fitting procedure and language-specific constraints described  in Appendix~\ref{sec:reproduction-specification}.
The composition-aware extension additionally accounts for the family and script composition of the training datasets.
It reduces mean RMSE by 0.000652 relative to the otherwise matched law without composition.

\begin{table*}[!t]
\centering\small
\begin{tabular}{lrrrr}
\toprule
Scaling law & Largest size & $E=32$ & $K=200$ & Mean \\
\midrule
Size and target-language tokens & 0.07250 & 0.11402 & 0.09994 & 0.09549 \\
+ Expert-dependent scaling & 0.05099 & 0.03989 & 0.06000 & 0.05029 \\
+ Size--data coupling & \textbf{0.03512} & 0.03381 & 0.06370 & 0.04421 \\
+ Language coverage & 0.03533 & 0.03049 & 0.04507 & 0.03696 \\
+ Language composition (composition-aware extension) & 0.03516 & \textbf{0.03022} & \textbf{0.04355} & \textbf{0.03631} \\
\midrule
Core joint law (without size--data coupling) & 0.05859 & 0.03830 & 0.05511 & 0.05067 \\
Composition-aware extension without size--data coupling & 0.05795 & 0.03790 & 0.05304 & 0.04963 \\
\bottomrule
\end{tabular}
\caption{Prediction RMSE for scaling law ablations across the three held-out configurations.}
\label{tab:validation-ablations}
\end{table*}

The Expert-dependent scaling row in Table~\ref{tab:validation-ablations} is an internal control obtained by setting $\phi(E)=\psi(E)=0$ in the core joint law (Equation~\ref{eq:joint-core}).
It therefore retains training tokens per language and expert count, but removes the language-coverage terms.
This control is distinct from the published-form JointMoE adaptation.
Adding language coverage to the expert-dependent control reduces $K=200$ RMSE from 0.06000 to 0.05511, while mean RMSE increases slightly from 0.05029 to 0.05067.
The control's all-grid fit obtains RMSE 0.27699 on the larger-model evaluation, compared with 0.27880 for the core joint law.

\paragraph{Ablating the core expert--language coverage interaction.}
\label{sec:interaction-ablation}
We test whether the coverage components $\phi$ and $\psi$ should depend on the number of experts. 
Table~\ref{tab:expert-coverage-ablation} compares variants in which neither, either, or both exponents depend on $E$.
Allowing both to depend on the number of experts gives the lowest mean RMSE, supporting the expert--language coverage interaction used in the core law.

\begin{table}[!t]
    \centering\small
    \begin{tabular*}{\columnwidth}{@{\extracolsep{\fill}}lr@{}}
        \toprule
        Coverage exponents & Mean RMSE \\
        \midrule
        Independent of expert count & 0.052362 \\
        Expert-dependent model-limited term & 0.051170 \\
        Expert-dependent data-limited term & 0.050980 \\
        Both expert-dependent & 0.050668 \\
        \bottomrule
    \end{tabular*}
    \caption{Mean Prediction RMSE across the three held-out configurations for different expert-dependent language coverage terms.}
    \label{tab:expert-coverage-ablation}
\end{table}

\begin{table}[!t]
    \centering\small
    \begin{tabular*}{\columnwidth}{@{\extracolsep{\fill}}lrr@{}}
        \toprule
        Evaluation & 14 coeffs. & 12 coeffs. \\
        \midrule
        Largest $N$ & \textbf{0.02912} & 0.02921 \\
        $E=32$ & 0.02587 & \textbf{0.02576} \\
        $K=200$ & 0.04496 & \textbf{0.04461} \\
        Mean RMSE & 0.03332 & \textbf{0.03319} \\
        Larger model & \textbf{0.17280} & 0.17298 \\
        \bottomrule
    \end{tabular*}
    \caption{Prediction RMSE for the 14-coefficient and selected 12-coefficient repetition-aware scaling laws across the held-out configurations and larger-model evaluation.}
    \label{tab:coverage-complexity}
\end{table}
We also compare repetition-aware laws with expert-dependent coverage exponents (14 coefficients) and fixed coverage exponents (12 coefficients).
We select the simpler 12-coefficient law because it achieves comparable predictive accuracy (Table~\ref{tab:coverage-complexity}).

\paragraph{Composition-aware extension.}
Motivated by evidence that cross-lingual transfer depends on language family and script \citep{longpre_2026_atlas}, we extend the scaling law to account for the language composition of the training data.
For target language $t$, we divide the non-target training tokens into tokens from languages in the same family, $D_{f,t}$, languages sharing the same script but not the same family, $D_{s,t}$, and all remaining languages, $D_{o,t}$.

We model the effects of language composition in different ways in the data- and model-limited terms.
Following prior work, we treat tokens from languages of the same family or the same script as potentially beneficial for the target language \citep{longpre_2026_atlas}, and model their interaction in the data-limited term.
Conversely, we model the remaining languages that are neither of the same family nor of the same script separately in the model-limited term, and thus allow unrelated-language data to capture additional capacity pressure.

We define
\begin{align}
 D_{\mathrm{comp},t}&=D_t
 \left(1+\frac{D_{f,t}}{D_t}\right)^{\eta_f}
 \left(1+\frac{D_{s,t}}{D_t}\right)^{\eta_s},
 \label{eq:composition-exposure}\\
 c_t&=\left(1+\frac{D_{o,t}}
 {D_t+D_{f,t}+D_{s,t}}\right)^{\gamma_o}.
 \label{eq:composition-capacity}
\end{align}
Here, $D_{\mathrm{comp},t}$ is the composition-adjusted amount of data for language $t$ and $c_t$ is the composition factor applied to the model-limited term.
The fitted coefficients $\eta_f$, $\eta_s$, and $\gamma_o$ control the strength of these composition effects.

Using these quantities, the composition-aware scaling law models $L_t(N,D_t,E,K)$ as
\begin{equation}
    \begin{aligned}
         L_{\infty,t} {}+
         \frac{A_tE^\delta K^{\phi(E)}c_t}   {N^{\alpha(E,K)}} +
         \frac{B_tE^\omega K^{\psi(E)}N^\zeta}  {D_{\mathrm{comp},t}^{\beta(E)}}.
    \end{aligned}
    \label{eq:coverage-coupled}
\end{equation}
Appendix~\ref{sec:composition-study} evaluates this law on alternative language mixtures, including its accuracy in predicting loss differences between mixtures (Table~\ref{tab:composition-prediction})

\paragraph{Repetition-aware controls.}
We compare two repetition-aware variants of the 14-coefficient law that vary in how model capacity affects the repetition discount.
In addition to the repetition-aware law presented in Section~\ref{sec:selected-law-specification}, we test a variant where the repetition parameter depends on $N$ rather than $P$, with $r=\lambda_0 N^\rho$.
Using $P$ gives lower RMSE across the primary holdouts from 0.03533 to 0.03332 and reduces larger-model RMSE from 0.19054 to 0.17280.
We therefore use stored capacity for the repetition discount in our selected law.

\paragraph{Baseline comparisons.}
We refit published scaling law baselines on the same structured validation splits.
Table~\ref{tab:joint-story-validation} reports Chinchilla and JointMoE fitted using target-language tokens, JointMoE using total training tokens, and adaptive ATLAS using either active or stored parameter counts.
ATLAS was originally developed for dense models, where active and total parameters coincide.
For our MoE models, we use active parameters $N$ as its standard model-size input and additionally evaluate total stored parameters $P$.
For completeness, Table ~\ref{tab:appendix-baselines} additionally reports Chinchilla using total training tokens, the ATLAS count/capacity law, and an adaptation of the scaling law of \citet{he_2025_multilingual_scaling}.
These baselines are adaptations to our data and evaluation protocol.


\begin{table*}[!t]
    \centering\small
    \begin{tabular}{lrrrr}
        \toprule
        Predictor & $N \approx 590$M & $E=32$ & $K=200$ & Mean \\
        \midrule
        Chinchilla, total tokens & 0.10630 & 0.13127 & 0.22741 & 0.15499 \\
        ATLAS count/capacity & 0.07130 & 0.11321 & 0.63113 & 0.27188 \\
        He et al. adaptation & 0.07312 & 0.11414 & 1.54303 & 0.57676 \\
        \midrule
        Composition-aware law & 0.03516 & 0.03022 & 0.04355 & 0.03631 \\
        \bottomrule
    \end{tabular}
    \caption{Prediction RMSE for additional baselines and the composition-aware law across the three held-out configurations.}
    \label{tab:appendix-baselines}
\end{table*}

\paragraph{Shared-coefficient controls.}
\label{sec:pooled-controls}
Some baseline scaling laws were not designed to jointly model scaling across many languages.
To give these baselines flexibility in a multilingual setting, we first fit their coefficients separately for each language.
Because our joint law instead shares its main scaling coefficients across languages, we also refit JointMoE and adaptive ATLAS with coefficients shared across languages.
This tests whether the advantage of our law comes simply from pooling information across languages or from its additional modeling structure.
The shared-coefficient versions still perform worse than our joint law, with mean RMSEs of 0.12895 for JointMoE ($D$) and 0.09549/0.08233 for the active/stored-parameter ATLAS variants.
This suggests that coefficient sharing alone does not explain the improvement.


\subsection{Parameterization of the number of experts}
\label{sec:expert-count-conventions}
In the selected law, only $\alpha$ and $\beta$ depend on $E$. 
We also evaluate a more flexible 14-coefficient variant in which $\phi$ and $\psi$ depend on $E$.
For each expert-dependent exponent, we use
\begin{align}
    q(E)=a_q+b_q\ln E.
\end{align}
This assumes that the exponent changes linearly with the logarithm of the number of experts, with slope $b_q$.
For instance, doubling the number of experts changes $q(E)$ by the same amount, $b_q\ln 2$.

We retain this log-linear assumption and equivalently express the function through its values at the smallest and largest numbers of experts in our study. 
Let $q_1=q(1)$ and $q_{32}=q(32)$. 
Since $\ln(1)=0$,
\[ q_1=a_q. \]
At $E=32$,
\[ q_{32}=q_1+b_q\ln 32, \]
which gives
\[ b_q=\frac{q_{32}-q_1}{\ln 32}. \]
Under the assumption that the exponent changes log-linearly with $E$, we can therefore equivalently write each exponent as
\begin{align}
    q(E) = q_1+(q_{32}-q_1)\frac{\ln E}{\ln 32}.
    \label{eq:expert-exponents-simple}
\end{align}

JointMoE follows \citet{clark_2022_unified_scaling} in introducing a transformed number of experts $\hat E$ to represent reduced efficiency at small numbers of experts and saturation at large numbers of experts:
\begin{equation}
    \frac{1}{\hat E} =
        \frac{1}{E-1+(1/E_{\mathrm{start}}-1/E_{\mathrm{max}})^{-1}}   + 
        \frac{1}{E_{\mathrm{max}}},
\end{equation}
The expert-dependent exponents are then modeled as
\begin{equation}
    q(E)=a_q+b_q\ln (\hat E(E)).
\end{equation}
This allows the effect of increasing the number of experts to vary across the range of $E$.
In particular, the transformation can represent reduced efficiency at small numbers of experts as well as saturation at large numbers of experts, as demonstrated by \citet{clark_2022_unified_scaling} and \citet{ludziejewski_2025_joint_moe}.
The parameters $E_{\mathrm{start}}$ and $E_{\mathrm{max}}$ control the shape of this transformation and thus introduce additional flexibility.
However they also introduce two additional parameters that are not required by Equation~\ref{eq:expert-exponents-simple}.

Since the joint law already contains several shared coefficients, we test whether the additional flexibility provided by the transformed number of experts improves predictive performance.
This sensitivity comparison uses the 14-coefficient repetition-aware law.
\begin{table}[!t]
    \centering\small
    \setlength{\tabcolsep}{5pt}
    \begin{tabular*}{\columnwidth}{@{\extracolsep{\fill}}lrr@{}}
        \toprule
        Evaluation & $E$ & $\hat E$ \\
        \midrule
        Largest size & 0.02912 & 0.02848 \\
        $E=32$ & 0.02587 & 0.02613 \\
        $K=200$ & 0.04496 & 0.04433 \\
        Primary mean & 0.03332 & 0.03298 \\
        \midrule
        $E=4$, full grid interpolation & 0.02680 & 0.02211 \\
        $E=16$, full grid interpolation & 0.02357 & 0.02329 \\
        $E=16$, restricted interpolation & 0.03455 & 0.07197 \\ 
        Larger model & 0.17280 & 0.17198 \\
        \bottomrule
    \end{tabular*}
    \caption{Prediction RMSE for the 14-coefficient, repetition-aware law with the raw expert count $E$ or the transformed count $\hat E$ across different evaluation settings.}
    \label{tab:expert-transform}
\end{table}
Across the three main validation splits, the two parameterizations give very similar prediction errors, with mean RMSE changing from $0.03332$ to
$0.03298$ (Table~\ref{tab:expert-transform}).
Additional evaluations focused on validation splits on the number of experts $E$ show some sensitivity, but no consistent advantage across settings.
Given the limited number of evaluated values of $E$ and the two additional fitted parameters required by $\hat E$, we retain the simpler log-linear parameterization in the law.

\paragraph{Parameterization sensitivity.}
We additionally test whether language coverage should affect the training-data exponent $\beta$, rather than only the active-parameters exponent $\alpha$.
We compare language-coverage dependence in $\alpha$ alone, in $\beta$ alone, and in both exponents with either a shared or separate coefficient.
Under the 14-coefficient formulation, none of the alternatives improves mean RMSE (Table~\ref{tab:coverage-exponent-ablation}), supporting our choice to retain $\alpha(E,K)$ and $\beta(E)$.

\begin{table*}[!t]
    \centering\small
    \begin{tabular}{lrrrr}
        \toprule
        Variant & $N \approx 590$M & $E=32$ & $K=200$ & Mean \\
        \midrule
        Coverage in $\alpha$ only (14-coefficient reference) & 0.02912 & 0.02587 & 0.04496 & \textbf{0.03332} \\
        Coverage in $\beta$ only & 0.03033 & 0.02601 & 0.04545 & 0.03393 \\
        Coverage in both (shared $\gamma$) & 0.02995 & 0.02603 & 0.04547 & 0.03382 \\
        Coverage in both (separate $\gamma_N,\gamma_D$) & 0.02969 & 0.02587 & 0.04498 & 0.03351 \\
        \bottomrule
    \end{tabular}
    \caption{Prediction RMSE across three held-out configurations for different language-coverage dependencies in the exponents $\alpha$ and $\beta$.}
    \label{tab:coverage-exponent-ablation}
\end{table*}

\subsection{Prediction diagnostics}
\label{sec:contrast-prediction}

\paragraph{Variance explained.}
To complement absolute prediction error, we report how much variation in observed loss is explained by the scaling law.
Table~\ref{tab:validation-r2} reports $R^2$ using the same predictions and weights as the RMSE evaluation:
\begin{equation}
 R^2 = 1- \frac{\sum_i w_i(L_i-\widehat L_i)^2} {\sum_i w_i(L_i-\overline L_w)^2},
\end{equation}
Here $L_i$ is unigram-adjusted loss, $\overline L_w=\sum_i w_iL_i$, and $\sum_iw_i=1$.
Because $R^2$ also captures variation between languages, we treat it as a supplementary diagnostic rather than primary evidence of generalization to held-out configurations.

\begin{table*}[t]
\centering\small
\begin{tabular}{@{}lrrrr@{}}
\toprule
Scaling law & Largest $N$ & $E=32$ & $K=200$ & Larger model \\
\midrule
Chinchilla ($D$) & 0.9579 & 0.9402 & 0.7981 & 0.7484\\
Chinchilla ($D_t$) & 0.9810 & 0.9554 & 0.9596 & 0.7528\\
JointMoE ($D$) & 0.9702 & 0.9739 & 0.8285 & 0.7119\\
JointMoE ($D_t$) & 0.9933 & 0.9961 & 0.9847 & 0.7240\\
Adaptive ATLAS ($N$) & 0.9813 & 0.9558 & 0.9597 & 0.7540\\
Adaptive ATLAS ($P$) & 0.9804 & 0.9710 & 0.9747 & 0.7053\\
\midrule
Core joint law & 0.9872 & 0.9949 & 0.9881 & 0.6493\\
+ size--data + size--coverage & 0.9953 & 0.9968 & 0.9921 & 0.5498\\
\textbf{Repetition-aware joint law (ours)} & \textbf{0.9968} & \textbf{0.9977} & \textbf{0.9922} & \textbf{0.8650}\\
\bottomrule\end{tabular}
\caption{Weighted $R^2$ for loss predictions across three held-out configurations and the larger-model evaluation (higher is better).}
\label{tab:validation-r2}
\end{table*}

The repetition-aware law achieves the highest explained variance across all evaluated settings, including the larger-model evaluation.

\subsection{Larger-model evaluation}
\label{sec:large-model-stress-test}
To test extrapolation beyond the original training grid, we additionally train a model with 1.149B active parameters and 21.23B total stored parameters, using 32 experts and 200 languages, trained for 68.7B tokens.
The model has 26 layers, width 1,664, 26 attention heads, and expert FFN width 4,992.
It uses the same Top-1 routing, tokenizer, corpus mixture, optimizer, 2,048-token context length, batch schedule, warmup, and linear learning-rate decay to zero over the final 20\% of training as the primary grid.
The run uses seed 1234 and the prescribed peak learning rate $1.0017\times10^{-4}$.
Evaluation uses the same 200-language validation sets with equal language weighting.

We first assess how scaling laws without repetition-adjusted training tokens per language extrapolate to this larger and longer-trained model.
Using the original $E=32$ holdout fits, RMSE is 0.28831 for the core law, 0.33367 for the composition-aware extension, 0.24352 for the active-parameter ATLAS adaptation, and 0.24363 for Chinchilla ($D_t$).
For reference, fitting the core law to the complete original grid gives RMSE 0.27880.

Prediction errors are particularly large for languages with substantial training-data repetition.
For the 137 languages with $D_t/U_t>4$, the composition-aware extension achieves RMSE 0.39373, compared with 0.12700 for the 62 languages with $1<D_t/U_t\leq4$.
The high-reuse group contributes 95.4\% of total squared error.
This highlights the importance of accounting for diminishing returns from repeated data when extrapolating to longer training.

We then compare repetition-aware formulations fitted on all 26,300 observations from the original grid, excluding the larger model from coefficient fitting.
The final stored-capacity repetition law achieves RMSE 0.17298, compared with 0.19054 for the 14-coefficient $N$-dependent repetition variant and 0.23349 for the strongest evaluated baseline on this run, ATLAS ($N$) (Table~\ref{tab:large-model-stress-test}).


\begin{table}[!t]
    \centering\small
    \begin{tabular*}{\columnwidth}{@{\extracolsep{\fill}}lr@{}}
        \toprule
        Predictor & RMSE \\
        \midrule
        Composition-aware extension & 0.31541 \\
        \midrule
        \multicolumn{2}{@{}l}{\textit{Repetition-aware variants}} \\
        $N$-dependent repetition, $\psi(E)$ & 0.19054 \\
        $N$-dependent repetition, shared $\psi$ & 0.19114 \\
        Stored-capacity repetition (ours) & \textbf{0.17298} \\
        \bottomrule
    \end{tabular*}
    \caption{Prediction RMSE on the larger model for different scaling-law formulations, with coefficients fitted on the original training grid.}
    \label{tab:large-model-stress-test}
\end{table}

\section{Robustness of the Expert--Language Coverage Response}
\label{sec:appendix-seeds}

\subsection{Training seeds and learning rate}
\label{sec:appendix-lr}
We test whether the expert--language coverage interaction remains consistent across training seeds and under a different learning rate.

\begin{figure}[!t]
    \centering
    \includegraphics[width=\columnwidth]{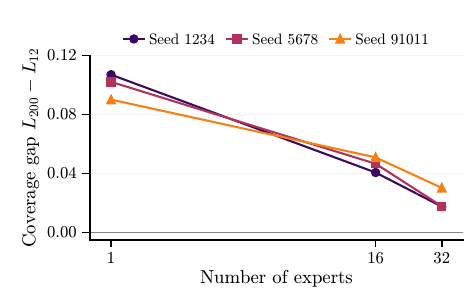}
    \caption{Differences in loss at matched $D_t$ with 190M active parameters across three seeds.}
    \label{fig:pressure-seeds}
\end{figure}

For the seed-replication experiments described in Appendix~\ref{par:seed-replication}, increasing the number of experts from 1 to 32 reduces the loss increase from expanding from 12 to 200 languages by 83.6\%, 82.6\%, and 66.4\% across three seeds, at matched training tokens per language (Figure~\ref{fig:pressure-seeds}).
Across all seeds, broader language coverage reduces loss at 2B tokens but increases it at 4B, 8B, and 16B tokens for both $E=16$ and $E=32$.
For $E=32$, the difference in loss ranges from $-0.0464$ to $-0.0307$ at 2B tokens, compared with $+0.0122$ to $+0.0283$ at 4B tokens.

The interaction is even stronger in a separate 64-configuration grid trained with a fixed learning rate of $2\times10^{-4}$.
Using the same comparison and weighting as in the primary analysis, increasing $E$ from 1 to 32 eliminates the mean loss increase from expanding from 12 to 200 languages, reducing it from 0.0708 to $-0.0003$.
This supports robustness of the interaction to the learning-rate choice.

\subsection{Token matching and interpolation}
\label{sec:appendix-pressure}

We detail how the differences in loss at matched $D_t$ in Section~\ref{sec:result-pressure} are computed and assess their robustness to interpolation choices.
We first determine each language's training token count $D_t$ from realized token counts.
For each language, active model size $N$, and expert count $E$, we subtract the $K=12$ loss, interpolated in log training tokens per language, from the $K=200$ loss at the same token count.

\begin{figure}[!t]
    \centering
    \includegraphics[width=\columnwidth]{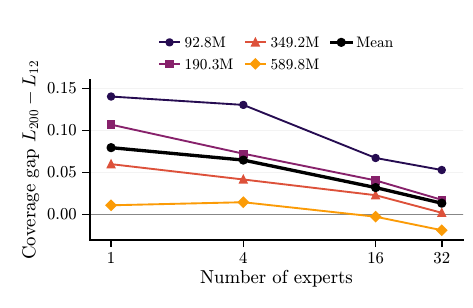}
    \caption{Mean loss difference for the same 12 languages when trained in a 200-language mixture versus a 12-language mixture, at matched training tokens per language, across four active model sizes. The black line shows the equally weighted mean across model sizes.}
    \label{fig:pressure}
\end{figure}

We only use overlapping ranges of training tokens per language, yielding 204 comparisons per expert count: 36, 48, 60, and 60 across the four sizes.
We average the differences in loss within each model size and expert count, then weight the four sizes equally (Figure~\ref{fig:pressure}).
As a robustness check, excluding each of the 12 languages in turn gives reductions in the loss increase from expanding from 12 to 200 languages from $E=1$ to $E=32$ between 77.8\% and 89.3\%.

For smoothed curves, we fit cubic B-splines in log token count, constrain them to be nonincreasing, and penalize curvature.
The smoothing strength is selected by held-out interior endpoints on the common-12 aggregate curves.
Sensitivity checks using neighboring strengths, an unconstrained smoother, PCHIP, and log-linear interpolation preserve the main conclusions.

\subsection{Sensitivity to model size and training duration}
\label{sec:capacity-duration-checks}

We examine how the expert--coverage interaction varies with active model size, training duration, and what is held fixed in the comparison.

At a $K=200$ training horizon of 4B tokens, increasing $E$ from 1 to 32 reduces the difference in loss at matched $D_t$ at both the smallest and largest values of $N$.
For the largest model, this reduction turns a slight loss increase into a benefit (Figure~\ref{fig:joint-capacity-response}a).

At equal total training tokens and 589.8M active parameters, the $E=1 \rightarrow 32$ loss advantage is 0.1680 for $K=12$ and 0.1618 for $K=200$ at 4B tokens (Figure~\ref{fig:joint-capacity-response}b).
At 32B tokens, the corresponding gains are 0.1098 and 0.1960.
The relative benefit of expert capacity therefore depends on language coverage and training duration.

\subsection{Other nested language expansions}
\label{sec:nested-pooling-audit}

We next test whether additional experts also reduce the loss increase for language expansions other than $K=12\to K=200$.
Across all six nested expansions, the mean loss difference is smaller with 32 experts than with a dense model in all 23 supported model-size and expansion comparisons, with a minimum reduction of 0.00655.
Table~\ref{tab:nested-matched-exposure} summarizes the nested transitions.
English-only comparisons have less overlapping ranges of training tokens per language, especially for $K=1\to200$.
We therefore use these analyses as complementary checks of the interaction across different language-coverage expansions.

\begin{table}[!t]
    \centering\small
    \setlength{\tabcolsep}{3pt}
    \begin{tabular}{@{}lrrrr@{}}
        \toprule
        $K$ expansion & $1\to12$ & $12\to50$ & $50\to200$ & $12\to200$ \\
        \midrule
        $E=1$ & 0.1075 & 0.0233 & 0.0359 & 0.0796 \\
        $E=4$ & 0.0485 & 0.0212 & 0.0188 & 0.0648 \\
        $E=16$ & 0.0498 & 0.0074 & -0.0017 & 0.0321 \\
        $E=32$ & 0.0420 & -0.0032 & -0.0046 & 0.0135 \\
        $E1-E32$ & 0.0655 & 0.0265 & 0.0405 & 0.0660 \\
        \bottomrule
    \end{tabular}
    \caption{Mean loss differences for the shared languages when expanding between language mixtures, at matched training tokens per language. Negative values favor broader language coverage. The last row shows the reduction in loss difference from increasing the number of experts from 1 to 32.}
    \label{tab:nested-matched-exposure}
\end{table}

\subsection{Resource-group robustness}
\label{sec:resource-group-robustness}

\begin{figure}[!t]
    \centering
    \includegraphics[width=\columnwidth]{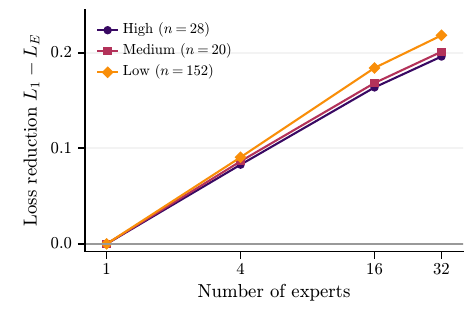}
    \caption{Mean loss reduction from increasing the number of experts relative to $E=1$, across high-, medium-, and low-resource languages at $K=200$.}
    \label{fig:resource-robustness}
\end{figure}

\paragraph{Absolute expert gains.}
At $K=200$, increasing from $E=1$ to $E=32$ lowers mean loss by 0.1960, 0.2010, and 0.2181 for high-, medium-, and low-resource languages, respectively.
Comparisons match language, $N$, and training tokens, then weight languages and values of $N$ equally.
Expert expansion therefore benefits all three resource groups, with the largest numerical gain in the low-resource group (Figure~\ref{fig:resource-robustness}).

\paragraph{Differences in loss at matched $D_t$.}
We test whether additional experts reduce the loss increase from expanding from $K=50$ to $K=200$ across language resource groups.
We group the 50 shared languages by resource level and compare losses at matched $D_t$ across all supported training horizons, weighting values of $N$ equally (Table~\ref{tab:resource-expansion}).
The loss difference is smaller with $E=32$ than with $E=1$ in every resource group.

\begin{table}[!t]
\centering\small
\begin{tabular}{@{}lrrrr@{}}
\toprule
Group & $n$ & $E=1$ diff. & $E=32$ diff. & Reduction \\
\midrule
High & 20 & $+0.0458$ & $+0.0057$ & 0.0401 \\
Medium & 6 & $+0.0226$ & $-0.0145$ & 0.0371 \\
Low & 16 & $+0.0222$ & $-0.0177$ & 0.0399 \\
Unknown & 8 & $+0.0484$ & $+0.0031$ & 0.0453 \\
\bottomrule
\end{tabular}
\caption{Mean loss differences between $K=200$ and $K=50$ for the same 50 languages, grouped by resource level and evaluated at matched training tokens per language. Negative values favor broader language coverage. Reduction denotes the decrease in loss difference from $E=1$ to $E=32$.}
\label{tab:resource-expansion}
\end{table}

\subsection{Alternative \texorpdfstring{$K=12$}{K=12} compositions}
\label{sec:composition-coverage}
We next test the robustness of the expert--coverage interaction to substantial changes in the $K=12$ language composition.
Besides the reference mixture, we train compact and diverse $K=12$ mixtures described in Appendix~\ref{sec:composition-study}.

For each mixture, we compare the seven languages shared across all conditions at $K=200$ horizons of 2, 4, 8, and 16B tokens, matching realized training tokens per language within observed support. 
Languages, horizons, and seeds are weighted equally.

\begin{table}[!t]
\centering\small
\setlength{\tabcolsep}{4pt}
\begin{tabular}{@{}lrrrr@{}}
\toprule
$K=12$ mixture & $E=1$ & $E=16$ & $E=32$ & Reduction \\
\midrule
Reference & 0.1024 & 0.0475 & 0.0234 & 77.2\% \\
Compact & 0.1051 & 0.0573 & 0.0246 & 76.6\% \\
Diverse & 0.0953 & 0.0355 & 0.0064 & 93.3\% \\
\bottomrule
\end{tabular}
\caption{Mean loss differences between $K=200$ and three alternative $K=12$ language mixtures, evaluated at matched training tokens per language. Lower values indicate smaller loss increases. Reduction denotes the decrease in loss difference from $E=1$ to $E=32$.}
\label{tab:composition-coverage}
\end{table}
Across all three mixtures, increasing $E$ from 1 to 32 reduces the loss increase from expanding from 12 to 200 languages by 76.6--93.3\%, confirming that the interaction is not specific to the original language mixture.

The same direction holds in every seed and at every evaluated horizon.
Alternative linear and monotone-cubic interpolation schemes also preserve the direction of the effect.
The reference value differs from the 83\% main-paper estimate because this analysis uses one value of $N$, seven shared languages, and three seeds.

\section{Language-Composition Analysis}
\label{sec:composition-study}
We further assess robustness to language composition by training compact and diverse mixtures at fixed $K=12$ and repeating the language-expansion comparison.

\paragraph{Design.}
We train two additional $K=12$ mixtures at 190M active parameters, including a compact mixture and a diverse mixture.
Each is trained with $E\in\{1,16,32\}$ across three seeds and six horizons from 0.5B to 16B tokens, yielding 18 additional training runs.

The two mixtures are constructed to differ in typological, genealogical, geographical, and script diversity while satisfying the same sampling and resource constraints.
Their memberships are listed in Table~\ref{tab:language-membership}. 
Seven languages are shared between them: Chinese, German, English, French, Portuguese, Russian, and Spanish.
We use these shared languages to evaluate loss prediction on changed mixtures and whether composition information improves prediction of the differences between them.

\paragraph{Loss prediction on changed mixtures.}
We first test whether the scaling laws predict loss accurately on alternative language mixtures without refitting.
We evaluate the final repetition-aware law and four published-law adaptations using their original-grid coefficients.
For each of three retained coefficient fits and each training seed, we compute RMSE over the seven shared languages, three expert counts, and two mixtures, then average across fits and seeds.

At 16B tokens, the repetition-aware law achieves RMSE 0.02649, compared with 0.10246 for the strongest evaluated baseline, JointMoE ($D$), a 74\% reduction (Table~\ref{tab:composition-loss}). 
This improvement holds across all three training seeds and all six evaluated training horizons.


\begin{table}[!t]
    \centering\small
    \begin{tabular*}{\columnwidth}{@{\extracolsep{\fill}}lr@{}}
        \toprule
        Predictor & Loss RMSE \\
        \midrule
        Chinchilla ($D_t$) & 0.12484 \\
        JointMoE ($D$) & 0.10246 \\
        ATLAS (active parameters) & 0.25974 \\
        ATLAS (stored parameters) & 0.22938 \\
        Repetition-aware joint law & \textbf{0.02649} \\
        \bottomrule
    \end{tabular*}
    \caption{Prediction RMSE on alternative language mixtures at 16B training tokens, without refitting the scaling laws (lower is better).}
    \label{tab:composition-loss}
\end{table}

\paragraph{Predicting changes in the language mixture.}
We next test whether explicitly modeling composition improves prediction of the loss difference between the diverse and compact mixtures.
We evaluate this difference on the seven shared languages at 16B tokens without refitting.
For each seed, RMSE weights languages and expert counts equally, then averages across the retained coefficient fits (Table~\ref{tab:composition-prediction}).

\begin{table}[!t]
    \centering\small
    \setlength{\tabcolsep}{3pt}
    \begin{tabular}{@{}lrrrr@{}}
        \toprule
        Predictor & S1 & S2 & S3 & Mean \\
        \midrule
        Composition-aware & \textbf{0.01335} & \textbf{0.01385} & \textbf{0.01180} & \textbf{0.01300} \\
        Without composition & 0.01771 & 0.01755 & 0.01597 & 0.01708 \\
        Without size--data & 0.01406 & 0.01434 & 0.01242 & 0.01361 \\
        Zero shift & 0.01730 & 0.01721 & 0.01555 & 0.01669 \\
        \bottomrule
    \end{tabular}
    \caption{Prediction RMSE for loss differences between the diverse and compact language mixtures at 16B tokens (lower is better). S1 to S3 denote training seeds.}
    \label{tab:composition-prediction}
\end{table}

The composition-aware extension reduces contrast RMSE from 0.01708 to 0.01300, a 23.9\% improvement over the otherwise matched law without composition.
It performs better in all nine fit--seed comparisons, with the same ordering across training horizons.

\paragraph{Composition and repetition capture different effects.}
The repetition-aware law achieves lower absolute-loss RMSE than the composition-aware extension (0.02649 versus 0.02898). Separately, adding composition to the non-repetition joint model reduces mixture-contrast RMSE from 0.01708 to 0.01300.
Thus, repetition and composition provide complementary predictive benefits.
Together with the consistent expert--coverage interaction across alternative mixtures (Appendix~\ref{sec:composition-coverage}), these results show that the main findings are not specific to the original $K=12$ mixture.

\section{Capacity Allocation}
\label{sec:appendix-allocation}
\label{sec:allocation-details}
All allocation analyses in this section use logical training compute, $F\approx6ND$, following standard MoE scaling-law accounting.
Measured hardware efficiency is reported separately in Appendix~\ref{sec:appendix-resource-systems}.

\subsection{Observed allocation crossover}
We examine when increasing training compute makes a configuration with more active parameters and fewer experts preferable.
The primary allocation result uses trained configurations within their observed token ranges and compares the same 12 languages across language-coverage levels.
Under a 0.4B total-parameter budget, we compare configurations with 92.8M active parameters and 16 experts against 190.3M active parameters and 4 experts, storing 375.9M and 381.4M parameters, respectively. 
The compute budget at which the preferred configuration switches occurs at $(1.70,2.56,2.75)\times10^{18}$ FLOPs for $K=12,50,200$, respectively, based on smoothed losses over the 12 shared languages.
The $K=200$/$K=12$ ratio is 1.62 (estimator range 1.61--1.67).
Broader language coverage therefore delays the switch toward more active parameters and fewer experts.

At $2\times10^{18}$ FLOPs, the 190.3M/E4 configuration is better by 0.0106 at $K=12$, whereas the 92.8M/E16 configuration is better by 0.0183 at $K=200$.
Log-linear and PCHIP interpolation preserve both decisions.

Under 0.7B and 2.2B total-parameter budgets, configurations with 32 experts remain preferred among the evaluated models across the shared observed range.
Under a total-parameter budget of approximately 5.1B, the configuration with 349M active parameters and 32 experts outperforms the one with 590M active parameters and 16 experts throughout their shared observed compute range.

\subsection{Fitted allocation surfaces}
\label{sec:fitted-allocation-details}
Figure~\ref{fig:fitted-allocation-language-shift} uses the composition-aware law to predict capacity allocation under fixed total-parameter budgets.
We use the composition-aware law because it makes the most accurate allocation decisions on the held-out $K=200$ configurations under a 0.4B total-parameter budget.
At the 5.1B-parameter budget, the predicted $E=32\to16$ switch moves from $4.51\times10^{19}$ FLOPs at $K=1$ to $1.22\times10^{20}$ FLOPs at $K=200$, a $2.70\times$ delay, with the latter extrapolated beyond observed training.
Other fitted formulations predict the same direction of change, although the magnitude varies.

\subsection{Preserving the original languages}
\begin{figure}[!t]
    \centering
    \includegraphics[width=\columnwidth]{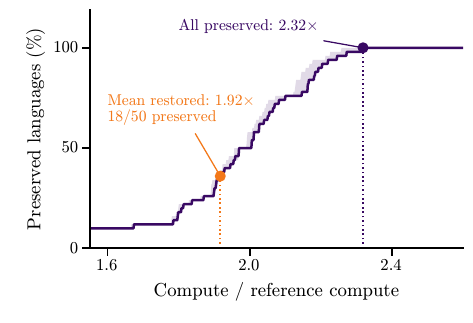}
    \caption{Fraction of the original 50 languages preserved after expansion to $K=200$. Shading shows interpolation sensitivity.}
    \label{fig:allocation-preservation-detail}
\end{figure}

We examine how much additional compute is needed to preserve the original languages when expanding language coverage.
Within observed training configurations, we identify those that match the original training tokens per language, restore mean loss, or preserve every original language.
For $K=50\to200$, a $2\times10^{18}$-FLOP reference and 2.2B total-parameter budget require approximately $1.75\times$, $1.92\times$, and $2.32\times$ compute, respectively (Figure~\ref{fig:allocation-preservation-detail}).
Restoring mean loss preserves only 18 of the original 50 languages, while 32 still experience higher loss, including both high- and low-resource languages.
Across interpolators, mean-preserving budgets range from 1.904 to 1.918 and all-language budgets from 2.258 to 2.319.

\begin{figure}[!t]
\centering
\includegraphics[width=\columnwidth]{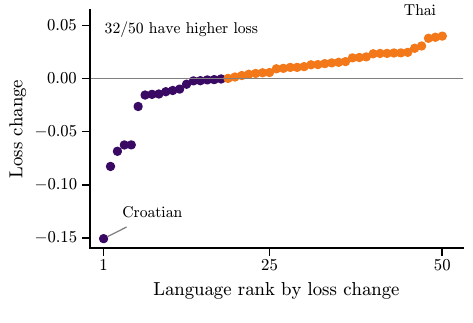}
\caption{Loss changes for the original 50 languages after expanding to $K=200$ at $1.92\times$ the reference compute, where their mean loss is preserved. Languages are sorted by loss change. Orange indicates higher loss, while purple indicates unchanged or lower loss.}
\label{fig:preservation-distribution}
\end{figure}

\section{External Evaluation}
\label{sec:appendix-benchmarks}

\subsection{Tasks and scoring}

We evaluate 14 multilingual benchmarks using the final checkpoints of all 64 models from the main experiment and 18 additional checkpoints from the seed-replication experiments (Appendix~\ref{par:seed-replication}).
Table~\ref{tab:benchmark-coverage} provides an overview of the benchmarks, grouped by task, along with their language coverage and number of evaluation examples.
All models are evaluated on the same examples for each benchmark.
Models of the same size $N$ are trained with the same total number of tokens and computational budget, but differ in training tokens per language. 
Therefore, the interaction magnitudes are not directly comparable to the 83\% estimate at matched $D_t$ in Section~\ref{sec:result-pressure}.

For multiple-choice evaluation, we use Belebele, Global MMLU, Okapi HellaSwag, XStoryCloze, XCOPA, Uhura ARC-Easy, and a European ARC-Easy release \citep{bandarkar_2024_belebele,singh_2025_global_mmlu,lai_2023_okapi,lin_2022_few_shot,ponti_2020_xcopa,bayes_2026_uhura,clark_2018_arc}.
These benchmarks are evaluated zero-shot using character-normalized answer likelihoods with lm-evaluation-harness \citep{gao_2021_evaluation_framework}.
For X-CSQA, we instead use PMI-adjusted answer likelihoods \citep{lin_2021_common_sense}.
MultiBLiMP and XCOMPS evaluate grammatical and conceptual knowledge, respectively, using minimal-pair accuracy \citep{jumelet_2026_multiblimp,he_2025_xcomps}.
SIB-200 and Taxi1500 evaluate text classification using three fixed English demonstrations, with accuracy, macro-F1, and majority-class baselines \citep{adelani_2024_sib200,ma_2025_taxi1500}.
Finally, we use FLORES-200 and NTREX-128 to measure monolingual text likelihood rather than translation quality \citep{nllb_team_2022_no_language_left_behind,federmann_2022_ntrex}.
We report negative log likelihood per token and bits per UTF-8 byte.

\begin{table}[!t]
\centering
\small
\setlength{\tabcolsep}{3pt}
\renewcommand{\arraystretch}{1.06}
\begin{tabular*}{\columnwidth}{@{\extracolsep{\fill}}lrr}
\toprule
\textbf{Benchmark} & \textbf{\# Lang.} & \textbf{Examples per language} \\
\midrule

\rowcolor{black!10}\multicolumn{3}{l}{\strut\textit{Reading comprehension}} \\
Belebele & 122 & 900 \\

\addlinespace[3pt]
\rowcolor{black!10}\multicolumn{3}{l}{\strut\textit{Knowledge and science question answering}} \\
Global MMLU & 42 & 14,035--14,042 \\
ARC / Uhura & 7 & 300--2,376 \\
ARC / EU20 & 20 & 2,367--2,376 \\

\addlinespace[3pt]
\rowcolor{black!10}\multicolumn{3}{l}{\strut\textit{Commonsense and narrative reasoning}} \\
Okapi HellaSwag & 32 & 8,412--10,042 \\
XCOPA & 12 & 500 \\
XStoryCloze & 11 & 1,511 \\
X-CSQA & 16 & 986--1,000 \\

\addlinespace[3pt]
\rowcolor{black!10}\multicolumn{3}{l}{\strut\textit{Grammatical and conceptual knowledge}} \\
MultiBLiMP & 61 & 7--4,483 \\
XCOMPS & 17 & 14,138--14,294 \\

\addlinespace[3pt]
\rowcolor{black!10}\multicolumn{3}{l}{\strut\textit{Text classification}} \\
SIB-200 & 205 & 204 \\
Taxi1500 & 877 & 1--111 \\

\addlinespace[3pt]
\rowcolor{black!10}\multicolumn{3}{l}{\strut\textit{Monolingual text likelihood}} \\
FLORES-200 & 204 & 1,012 \\
NTREX-128 & 128 & 123 \\

\bottomrule
\end{tabular*}
\caption{Number of language varieties and evaluation examples per variety for each benchmark. NTREX-128 counts documents, while other benchmarks count individual examples.}
\label{tab:benchmark-coverage}
\end{table}

\subsection{Expert gains across language coverage}
Table~\ref{tab:benchmark-scores} reports results for all 64 configurations, comparing expert counts at fixed active parameters and total training tokens within each language-coverage setting.
For each language-coverage setting, we average benchmark scores across its training languages with equal language weights.
Since the evaluated languages differ across $K$, these comparisons show expert gains within each setting.

\begin{table*}[!p]
\centering
\fontsize{8}{9}\selectfont
\setlength{\tabcolsep}{1pt}
\renewcommand{\arraystretch}{0.95}
\begin{tabular*}{\textwidth}{@{\extracolsep{\fill}}lr*{4}{r}@{\hspace{5pt}}*{4}{r}@{\hspace{5pt}}*{4}{r}@{\hspace{5pt}}*{4}{r}@{}}
\toprule
 & & \multicolumn{4}{c}{$K=1$} & \multicolumn{4}{c}{$K=12$} & \multicolumn{4}{c}{$K=50$} & \multicolumn{4}{c}{$K=200$} \\
\cmidrule(lr){3-6}\cmidrule(lr){7-10}\cmidrule(lr){11-14}\cmidrule(lr){15-18}
Benchmark & $N$ (M) & $E=1$ & $4$ & $16$ & $32$ & $E=1$ & $4$ & $16$ & $32$ & $E=1$ & $4$ & $16$ & $32$ & $E=1$ & $4$ & $16$ & $32$ \\
\midrule
\rowcolor{black!10}\multicolumn{18}{l}{\strut\textit{Reading comprehension}} \\
Belebele & 93 & \textbf{28.44} & 28.00 & 28.11 & 28.22 & \textbf{27.42} & 27.23 & 27.37 & 27.12 & 27.27 & 26.94 & 27.11 & \textbf{27.28} & 26.76 & \textbf{27.05} & 26.69 & 26.66 \\
 & 190 & 30.00 & 29.89 & \textbf{31.44} & 31.22 & 28.20 & 28.40 & 28.71 & \textbf{28.98} & 27.44 & 27.65 & \textbf{27.67} & 27.57 & 26.96 & 27.09 & 27.00 & \textbf{27.15} \\
 & 349 & 30.00 & 31.11 & \textbf{32.67} & 32.00 & 28.88 & \textbf{30.00} & 29.91 & 29.53 & 28.21 & \textbf{28.61} & 28.35 & 28.29 & \textbf{27.82} & 27.77 & 27.51 & 27.54 \\
 & 590 & 32.00 & 31.44 & 31.56 & \textbf{33.33} & 29.79 & \textbf{30.08} & 30.05 & 29.81 & \textbf{28.78} & 28.52 & 28.44 & 28.75 & 27.93 & \textbf{28.02} & 27.74 & 27.75 \\
\rowcolor{black!10}\multicolumn{18}{l}{\strut\textit{Knowledge and science QA}} \\
Global MMLU & 93 & 25.85 & 26.10 & 26.58 & \textbf{26.95} & 25.58 & \textbf{25.71} & 25.67 & 25.65 & \textbf{25.67} & \textbf{25.67} & 25.58 & 25.59 & \textbf{25.63} & 25.61 & 25.57 & 25.59 \\
 & 190 & 26.51 & 26.94 & 27.29 & \textbf{27.32} & 25.61 & 25.58 & 25.76 & \textbf{25.78} & 25.59 & 25.52 & 25.58 & \textbf{25.65} & 25.56 & \textbf{25.57} & 25.52 & 25.56 \\
 & 349 & 26.81 & 27.57 & 28.54 & \textbf{28.76} & 25.81 & 25.85 & \textbf{26.25} & \textbf{26.25} & 25.66 & 25.73 & 25.77 & \textbf{25.84} & 25.57 & 25.62 & 25.66 & \textbf{25.68} \\
 & 590 & 27.72 & 28.36 & 29.05 & \textbf{30.07} & 26.11 & 26.08 & 26.22 & \textbf{26.32} & 25.73 & 25.81 & 25.86 & \textbf{26.01} & 25.59 & \textbf{25.76} & \textbf{25.76} & 25.71 \\
ARC / Uhura & 93 & 35.23 & 35.61 & 35.82 & \textbf{37.92} & 32.20 & 32.87 & 32.66 & \textbf{33.50} & \textbf{31.35} & 30.24 & 30.53 & 30.83 & 28.87 & \textbf{29.65} & 29.40 & 29.42 \\
 & 190 & 38.47 & 38.89 & 40.53 & \textbf{41.75} & 34.30 & 34.76 & 35.94 & \textbf{36.11} & 30.34 & 30.28 & \textbf{31.11} & 30.45 & 30.56 & 29.92 & \textbf{30.64} & 30.55 \\
 & 349 & 42.93 & 43.60 & \textbf{46.93} & 46.68 & 35.77 & 36.95 & \textbf{39.31} & 39.23 & 31.57 & 31.57 & \textbf{32.70} & 32.00 & 30.90 & 30.72 & \textbf{31.22} & 30.98 \\
 & 590 & 44.02 & 45.50 & 49.07 & \textbf{50.55} & 37.50 & 38.97 & 40.32 & \textbf{40.87} & 32.32 & 32.12 & 31.81 & \textbf{32.34} & 30.27 & 31.96 & 31.09 & \textbf{32.30} \\
ARC / EU20 & 93 & -- & -- & -- & -- & 29.35 & 29.81 & 29.87 & \textbf{30.22} & 28.07 & 28.82 & 29.45 & \textbf{29.50} & 28.11 & 28.17 & 28.76 & \textbf{28.99} \\
 & 190 & -- & -- & -- & -- & 30.79 & 31.38 & 32.51 & \textbf{32.67} & 29.81 & 30.28 & 30.79 & \textbf{31.87} & 29.32 & 29.59 & 30.50 & \textbf{30.59} \\
 & 349 & -- & -- & -- & -- & 32.79 & 33.98 & 35.04 & \textbf{35.11} & 31.59 & 32.38 & 33.11 & \textbf{33.58} & 30.61 & 31.26 & 31.81 & \textbf{32.02} \\
 & 590 & -- & -- & -- & -- & 33.62 & 35.19 & 35.61 & \textbf{35.74} & 32.89 & 33.44 & 34.31 & \textbf{34.61} & 31.65 & 32.30 & 32.54 & \textbf{32.90} \\
\rowcolor{black!10}\multicolumn{18}{l}{\strut\textit{Commonsense and narrative reasoning}} \\
HellaSwag & 93 & 29.03 & 29.57 & 31.56 & \textbf{32.92} & 27.71 & 27.82 & 28.01 & \textbf{28.22} & 27.51 & 27.48 & \textbf{27.66} & 27.56 & \textbf{27.50} & 27.45 & 27.38 & 27.42 \\
 & 190 & 34.02 & 36.00 & 40.48 & \textbf{42.32} & 29.05 & 29.65 & 30.44 & \textbf{30.74} & 28.26 & 28.50 & 28.82 & \textbf{29.00} & 27.96 & 27.94 & 28.14 & \textbf{28.35} \\
 & 349 & 40.81 & 44.06 & 50.57 & \textbf{51.85} & 31.28 & 32.35 & 34.10 & \textbf{34.39} & 29.96 & 30.36 & 31.23 & \textbf{31.61} & 28.78 & 29.26 & 29.61 & \textbf{29.85} \\
 & 590 & 45.43 & 48.94 & 55.49 & \textbf{57.27} & 32.98 & 34.23 & \textbf{35.75} & 35.73 & 31.03 & 31.74 & 32.35 & \textbf{32.46} & 29.68 & 30.06 & 30.36 & \textbf{30.65} \\
XCOPA & 93 & 55.20 & 55.80 & 57.00 & \textbf{60.00} & 52.65 & 52.25 & 52.90 & \textbf{53.45} & 51.40 & 51.32 & 51.25 & \textbf{51.90} & 51.07 & 50.69 & 50.67 & \textbf{51.11} \\
 & 190 & 60.00 & 61.80 & 64.20 & \textbf{67.20} & 53.85 & 54.20 & 54.25 & \textbf{55.80} & 51.90 & 52.33 & 52.95 & \textbf{53.33} & 50.51 & 51.16 & 51.27 & \textbf{52.09} \\
 & 349 & 65.60 & 67.80 & \textbf{70.40} & 70.00 & 54.80 & 56.75 & \textbf{56.90} & 56.70 & 54.60 & 54.62 & \textbf{55.00} & 54.77 & 52.31 & 52.31 & \textbf{52.36} & 52.29 \\
 & 590 & 68.40 & 70.60 & \textbf{74.00} & 72.80 & 56.60 & \textbf{59.85} & 58.30 & 58.35 & 54.67 & 54.60 & 55.65 & \textbf{55.90} & 52.62 & 52.58 & \textbf{53.00} & 52.78 \\
XStoryCloze & 93 & 58.50 & 59.76 & \textbf{61.28} & 60.29 & 55.65 & 55.99 & 56.35 & \textbf{56.61} & 55.32 & 55.54 & 55.22 & \textbf{55.76} & 55.21 & \textbf{55.27} & 55.04 & 54.85 \\
 & 190 & 61.81 & 62.74 & 65.06 & \textbf{65.92} & 57.13 & 57.11 & 58.23 & \textbf{58.29} & 55.95 & 56.32 & 56.68 & \textbf{56.95} & \textbf{56.12} & 55.57 & 55.68 & 55.80 \\
 & 349 & 65.45 & 66.25 & 68.83 & \textbf{69.36} & 58.80 & 59.46 & 60.77 & \textbf{60.85} & 57.86 & 57.70 & \textbf{58.85} & 58.61 & 57.00 & 57.11 & \textbf{57.49} & 57.38 \\
 & 590 & 67.44 & 69.03 & \textbf{72.40} & 72.27 & 60.19 & 61.13 & 61.26 & \textbf{61.65} & 58.68 & 58.74 & 59.01 & \textbf{59.09} & 57.56 & 57.82 & 57.88 & \textbf{58.13} \\
X-CSQA & 93 & 32.90 & 31.70 & \textbf{33.30} & 32.30 & 22.48 & \textbf{23.78} & 23.31 & 23.17 & 21.55 & 21.97 & \textbf{22.56} & 21.84 & 21.36 & 21.30 & 22.18 & \textbf{22.40} \\
 & 190 & \textbf{38.60} & 36.30 & 35.70 & 36.80 & 25.28 & 24.68 & 25.97 & \textbf{26.28} & 23.57 & 23.37 & \textbf{24.14} & 23.96 & 22.78 & 22.58 & 23.33 & \textbf{23.66} \\
 & 349 & 42.20 & 41.20 & \textbf{44.30} & 43.80 & 27.35 & 26.47 & \textbf{28.71} & 28.43 & 25.13 & 25.04 & 25.58 & \textbf{26.10} & 23.65 & 24.66 & 24.67 & \textbf{25.57} \\
 & 590 & 41.40 & 44.20 & \textbf{47.90} & 46.90 & 28.46 & 27.76 & 28.31 & \textbf{28.86} & 25.84 & 25.61 & 26.93 & \textbf{26.98} & 25.16 & 25.12 & 25.75 & \textbf{25.78} \\
\rowcolor{black!10}\multicolumn{18}{l}{\strut\textit{Grammatical and conceptual knowledge}} \\
MultiBLiMP & 93 & 96.36 & 95.84 & 96.75 & \textbf{97.14} & 90.97 & 91.49 & 92.83 & \textbf{92.95} & 87.53 & 88.39 & 90.25 & \textbf{90.64} & 81.99 & 82.33 & 84.16 & \textbf{85.21} \\
 & 190 & 98.05 & \textbf{98.31} & 97.53 & 97.92 & 94.30 & 94.94 & 95.52 & \textbf{95.69} & 92.03 & 92.05 & 93.27 & \textbf{93.83} & 86.17 & 87.17 & 88.47 & \textbf{89.13} \\
 & 349 & 97.79 & \textbf{98.05} & 97.14 & 97.66 & 95.68 & 95.98 & \textbf{96.57} & 96.46 & 93.84 & 93.71 & 94.36 & \textbf{94.63} & 89.04 & 89.57 & 90.50 & \textbf{91.14} \\
 & 590 & \textbf{98.57} & 98.05 & 98.31 & 98.05 & 96.20 & \textbf{96.73} & 96.47 & 96.15 & 94.57 & 94.56 & 94.43 & \textbf{94.61} & 90.35 & 91.07 & \textbf{91.44} & 91.10 \\
XCOMPS & 93 & -- & -- & -- & -- & 52.90 & 53.34 & 53.89 & \textbf{54.01} & 51.65 & 51.58 & \textbf{52.20} & 52.03 & 50.90 & 51.17 & \textbf{51.51} & 51.37 \\
 & 190 & -- & -- & -- & -- & 54.96 & 55.64 & \textbf{56.99} & 56.50 & 52.65 & 52.75 & 53.44 & \textbf{53.89} & 51.98 & 52.21 & 52.41 & \textbf{52.67} \\
 & 349 & -- & -- & -- & -- & 56.95 & 57.95 & 58.58 & \textbf{58.83} & 54.24 & 54.42 & 55.54 & \textbf{55.86} & 52.70 & 53.18 & 53.85 & \textbf{54.54} \\
 & 590 & -- & -- & -- & -- & 57.90 & 58.66 & 59.06 & \textbf{59.17} & 54.88 & 55.23 & \textbf{56.50} & 56.46 & 53.34 & 54.14 & 54.68 & \textbf{54.91} \\
\rowcolor{black!10}\multicolumn{18}{l}{\strut\textit{Text classification}} \\
SIB-200 & 93 & 27.94 & \textbf{28.43} & 19.61 & 21.57 & 26.43 & 23.12 & 23.33 & \textbf{27.61} & 25.72 & 25.75 & 25.29 & \textbf{25.82} & 22.50 & 21.97 & 18.92 & \textbf{25.39} \\
 & 190 & 24.51 & 20.10 & 19.61 & \textbf{25.98} & \textbf{27.08} & 20.79 & 26.27 & 20.51 & 20.52 & 22.48 & 24.77 & \textbf{25.45} & 22.15 & 24.74 & \textbf{24.91} & 22.90 \\
 & 349 & 23.04 & \textbf{26.47} & 19.61 & 17.16 & \textbf{23.61} & 22.71 & 17.16 & 21.28 & \textbf{22.19} & 19.25 & 19.91 & 21.67 & 24.42 & 18.49 & \textbf{27.18} & 25.32 \\
 & 590 & 30.88 & \textbf{34.31} & 14.22 & 14.71 & 22.55 & 20.34 & 19.12 & \textbf{24.88} & 21.45 & 21.38 & \textbf{23.24} & 21.16 & 24.40 & 17.19 & 21.45 & \textbf{25.13} \\
Taxi1500 & 93 & 9.91 & \textbf{28.83} & 27.03 & 18.92 & \textbf{24.40} & 6.68 & 18.39 & 22.15 & \textbf{22.70} & 14.42 & 10.77 & 20.39 & 10.38 & \textbf{17.78} & 16.68 & 16.62 \\
 & 190 & 17.12 & 17.12 & \textbf{18.02} & 14.41 & 6.91 & 12.69 & 11.49 & \textbf{16.97} & 14.78 & 6.41 & \textbf{25.80} & 21.73 & 11.93 & 7.02 & \textbf{19.73} & \textbf{19.73} \\
 & 349 & 16.22 & 17.12 & \textbf{23.42} & 18.92 & 18.39 & 16.44 & 16.74 & \textbf{22.15} & 14.91 & \textbf{17.30} & 17.00 & 17.10 & 14.72 & \textbf{16.86} & 15.18 & 14.81 \\
 & 590 & 12.61 & 14.41 & 20.72 & \textbf{21.62} & 15.39 & 13.59 & 17.94 & \textbf{18.54} & 14.42 & 14.23 & \textbf{17.44} & 13.86 & 17.22 & 18.68 & \textbf{25.45} & 19.61 \\
\rowcolor{black!10}\multicolumn{18}{l}{\strut\textit{Monolingual text likelihood}} \\
FLORES-200 & 93 & 1.247 & 1.230 & 1.192 & \textbf{1.185} & 1.270 & 1.235 & 1.192 & \textbf{1.180} & 1.462 & 1.425 & 1.371 & \textbf{1.352} & 1.642 & 1.591 & 1.534 & \textbf{1.498} \\
 & 190 & 1.159 & 1.134 & 1.098 & \textbf{1.086} & 1.158 & 1.124 & 1.092 & \textbf{1.086} & 1.333 & 1.296 & 1.259 & \textbf{1.240} & 1.480 & 1.423 & 1.372 & \textbf{1.351} \\
 & 349 & 1.098 & 1.077 & 1.047 & \textbf{1.035} & 1.088 & 1.060 & 1.040 & \textbf{1.036} & 1.240 & 1.215 & 1.193 & \textbf{1.188} & 1.364 & 1.323 & 1.282 & \textbf{1.275} \\
 & 590 & 1.073 & 1.052 & 1.021 & \textbf{1.015} & 1.061 & 1.039 & \textbf{1.034} & 1.049 & 1.202 & 1.186 & \textbf{1.179} & 1.189 & 1.311 & 1.281 & \textbf{1.259} & 1.266 \\
NTREX-128 & 93 & 1.083 & 1.051 & 1.012 & \textbf{0.993} & 1.099 & 1.057 & 1.025 & \textbf{1.011} & 1.153 & 1.117 & 1.076 & \textbf{1.062} & 1.251 & 1.216 & 1.168 & \textbf{1.142} \\
 & 190 & 0.969 & 0.941 & 0.890 & \textbf{0.878} & 0.984 & 0.945 & 0.913 & \textbf{0.906} & 1.034 & 0.996 & 0.968 & \textbf{0.954} & 1.118 & 1.076 & 1.039 & \textbf{1.023} \\
 & 349 & 0.891 & 0.862 & 0.819 & \textbf{0.810} & 0.912 & 0.878 & \textbf{0.852} & 0.853 & 0.949 & 0.920 & 0.898 & \textbf{0.895} & 1.019 & 0.983 & 0.960 & \textbf{0.952} \\
 & 590 & 0.855 & 0.826 & 0.788 & \textbf{0.780} & 0.879 & 0.851 & \textbf{0.843} & 0.856 & 0.913 & 0.890 & \textbf{0.882} & 0.888 & 0.974 & 0.950 & \textbf{0.938} & 0.940 \\
\bottomrule
\end{tabular*}
\caption{Benchmark scores across the 64 main training configurations. Scores are accuracy (\%, higher is better), except for FLORES and NTREX (bits/byte, lower is better). Each $K$ averages over its benchmark-supported training languages. Bold indicates the best score across expert counts at fixed $N$ and $K$. Dashes indicate unavailable evaluations.}
\label{tab:benchmark-scores}
\end{table*}

\subsection{Expert--coverage interaction on shared languages}
We additionally compare expert gains at $K=12$ and $K=200$ using the same benchmark-supported languages from the original 12-language training set.
We average varieties within each language and weight languages equally.
These comparisons hold active parameters and total training tokens fixed, but not training tokens per language.






FLORES and NTREX show larger $E=1\to32$ loss improvements at $K=200$ than at $K=12$ across all four active sizes and all three 190M seeds.
For the default seed, expert expansion reduces the loss increase from expanding from 12 to 200 languages by 22--57\% on FLORES and 20--62\% on NTREX.
The same interaction direction holds for both token- and byte-normalized loss.
MultiBLiMP shows larger grammaticality-accuracy gains at $K=200$ across the same configurations.

Downstream accuracy interactions are more task-dependent.
HellaSwag and XStoryCloze show smaller expert gains at $K=200$ across all four active sizes and three 190M seeds, while other benchmarks exhibit mixed effects.
For HellaSwag, this negative accuracy interaction is consistent across scoring normalizations.
Thus, the external likelihood results support the direction of the main loss finding, but downstream accuracy does not follow a uniform pattern.

\section{Routing Diagnostics}
\jd{work in progress}
\label{sec:appendix-routing}

Motivated by prior work on language-dependent expert routing in multilingual MoEs \citep{srivastava_2025_lola,bandarkar_2026_multilingual_routing}, we investigate whether routing patterns accompany the loss benefits of broader language coverage.

We collect Top-1 expert assignments at the final checkpoints of 60 training runs, excluding padding. We analyze the same 12 languages with equal weights across $K\in\{12,50,200\}$ and compare $E\in\{4,16,32\}$ within each model size $N$. Total training tokens are fixed at 8/16/32/32B for the four model sizes, respectively, but training tokens per language differ across $K$. We summarize language-dependent routing using language--expert mutual information (MI), within-language expert use using the entropy-based effective expert fraction, and cross-language routing similarity using pairwise divergence.

\paragraph{Findings.}
Broader language coverage does not consistently strengthen language-dependent routing. At $E=32$, increasing $K$ from 12 to 200 changes MI by $-0.0162$, $+0.1024$, $-0.0431$, and $-0.0440$ at 93M, 190M, 349M, and 590M, respectively.

The loss benefits of additional experts also lack a consistent counterpart in routing. At 190M, increasing $E$ from 16 to 32 yields an additional loss reduction at $K=200$ relative to $K=12$ of $0.0216$, $0.0284$, and $0.0260$ across three seeds. The corresponding MI interactions are $-0.0349$, $+0.0890$, and $-0.0753$. Thus, the loss interaction has a consistent direction across seeds, whereas the MI interaction does not.

Finally, cross-language routing distributions tend to be more similar in intermediate layers. In 29 of 36 configurations, mean pairwise routing divergence is lower in the middle third of layers than the average of the early and late thirds.

Overall, these diagnostics do not reveal a consistent association between stronger language-dependent routing and the observed loss benefits.
We leave a more detailed investigation of routing mechanisms to future work.

\section{Physical Training Costs}
\label{sec:appendix-resource-systems}

The main-paper allocation analyses use logical compute, $F\approx6ND$, following standard scaling-law practice and excluding systems overhead from sparse expert routing.
Across 100 comparisons matched by model size, language coverage, and training horizon, $E=4$, $16$, and $32$ require 1.17, 1.44, and 1.69 times as many GPU-hours as $E=1$, respectively (median).
Mean model FLOP utilization decreases from 35.7\% at $E=1$ to 22.0\% at $E=32$.
These measurements reflect our implementation rather than an inherent efficiency limit of MoE architectures.
Optimizing throughput across the wide range of configurations was beyond the scope of this study, and hardware-aware tuning may substantially improve MoE efficiency.

The 101 main-study runs, including evaluation and diagnostics, consumed approximately 5,900 H100 GPU-hours.
This excludes learning-rate calibration and earlier development runs.

\end{document}